\documentclass[lettersize,journal]{IEEEtran}
\usepackage{amsmath,amsfonts}
\usepackage{algorithmic}
\usepackage{algorithm}
\usepackage{array}
\usepackage[caption=false,font=normalsize,labelfont=sf,textfont=sf]{subfig}
\usepackage{textcomp}
\usepackage{stfloats}
\usepackage{url}
\usepackage{bm}
\usepackage{verbatim}
\usepackage{booktabs}
\usepackage{multirow}
\usepackage{graphicx}
\usepackage{cite}

\begin{document}

\title{POST: Prior–Observation Adversarial Learning of Spatio–Temporal Associations for Multivariate Time Series Anomaly Detection}

\author{Suofei Zhang, Yaxuan Zheng, and Haifeng Hu
\thanks{Suofei Zhang is with the School of Internet of Things, Nanjing University of Posts and Telecommunications, Nanjing 210003, China (e-mails: zhangsuofei@njupt.edu.cn).}
\thanks{Yaxuan Zheng, and Haifeng Hu are with the National Engineering Research Center of Communications and Networking, Nanjing University of Posts and Telecommunications, Nanjing 210003, China (e-mails: 1022010111@njupt.edu.cn; xfuwu@ieee.org; quan.zhou@njupt.edu.cn; huhf@njupt.edu.cn;).}
}

\markboth{Journal of \LaTeX\ Class Files,~Vol.~14, No.~8, August~2021}%
{Shell \MakeLowercase{\textit{et al.}}: A Sample Article Using IEEEtran.cls for IEEE Journals}

\IEEEpubid{0000--0000/00\$00.00~\copyright~2021 IEEE}

\maketitle 

\begin{abstract}
Existing Multivariate Time Series Anomaly Detection (MTSAD) frameworks increasingly rely on integrating Graph Neural Networks (GNNs) with sequence models to capture complex spatio-temporal dependencies.
However, less attention is paid to the spatial over-generalization problem, where unconstrained structural modeling indiscriminately reconstructs anomalies, inevitably degrading detection recall.
To tackle this problem, we propose a novel framework that unifies spatio-temporal modeling through a joint prior-observation adversarial learning paradigm.
In the spatial dimension, the model alternately learns adjacency matrices as structural prior and models the association discrepancy between prior and data-driven observation in a minimax manner during training.
Such adversarial optimization not only improves the model sensitivity for time-wise detection, but also enables the model to localize anomalies to specific channels.
To systematically evaluate this anomaly localization capability, we further construct a synthetic benchmark equipped with precise channel-wise annotations.
Extensive experiments across public datasets and our dedicated benchmark demonstrate that the proposed framework establishes a new state-of-the-art in both time-wise detection and spatial localization tasks. Our code, pre-trained models, and benchmark are publicly available at https://github.com/anocodetest1/POST.
\end{abstract}

\section{Introduction} 
\IEEEPARstart{M}{odern} infrastructures in diverse domains such as industrial manufacturing~\cite{7469060}, network security~\cite{10.1145/3447548.3467174}, and financial risk control~\cite{Cheng_Xiang_Shang_Zhang_Yang_Zhang_2020} are increasingly instrumented with a multitude of sensors and key performance indicators, generating massive Multivariate Time Series (MTS) data that encode critical information about system states. 
Anomaly Detection (AD), which aims to identify observations or segments that deviate from established normal patterns in MTS, is a crucial task for mitigating catastrophic failures and ensuring operational stability.
Real-world MTS often exhibit two primary properties: (1) high-dimensional complexity, characterized by intricate temporal dynamics and latent spatial coupling; (2) massive normal data with rare, diverse, and noise-sensitive anomalies.
In a nutshell, MTSAD is an open-set problem.
Particularly because the distribution of anomalies is inherently unbounded and unpredictable, unsupervised learning has become the prevailing standard in the community.
Under this paradigm, models characterize the normal operational manifold from routine data and utilize prediction or reconstruction deviations as proxies for anomaly scoring.
\IEEEpubidadjcol

The methodology of MTSAD has evolved from time-agnostic outlier detection to explicit temporal modeling.
Early unsupervised approaches~\cite{10.1145/342009.335388,hamilton2020time,Anderson1976TimeSeries2E,https://doi.org/10.1002/for.768} relying on distance, density, or one-class boundaries typically falter when confronted with high-dimensional data and long-span dependencies.
Consequently, deep learning has significantly prompted the development of MTSAD and established two prevailing trajectories: (1) prediction-based methods exploit Recurrent Neural Networks (RNNs)~\cite{s23249878,10.1145/3219819.3219845,10.1145/3292500.3330672} to capture transient dynamics and forecast future observations.
Anomalies are scored by the discrepancy between predictions and actual observations. 
(2) reconstruction-based methods exploit Autoencoders (AE) and Generative Adversarial Networks (GANs)~\cite{9618824} to project data into a constrained normal manifold and isolate anomalies via reconstruction residuals.

Recently, the Transformer, serving as a powerful AE backbone, has been introduced to MTSAD~\cite{xu2022anomaly,10.14778/3514061.3514067,FU2024119978,10.1016/j.neucom.2024.129024}.
Leveraging Multi-Head Self-Attention (MHSA)~\cite{NIPS2017_3f5ee243}, these methods explicitly model global temporal dependencies and achieve superior capacity in capturing complex patterns in long sequences.
Despite the powerful reconstruction ability of Transformers, a typical paradox in the unsupervised paradigm of MTSAD is non-negligible.
Without extra constraints, there exists no theoretical guarantee that the model will only reconstruct normal patterns rather than over-generalize and reconstruct anomalies as well.
To address this, the Anomaly Transformer (AT)~\cite{xu2022anomaly} introduces an explicit adversarial mechanism between the prior and the observation of the temporal association.
Equipped with adversarial optimization and the resulting quantitative association discrepancy, AT not only prevents model from unconstrained generalization, but also provides a direct metric of the temporal integrity of signal.

Despite these improvements, we argue that the exploration of latent sptatial topologies among sensors in MTS remains inadequate.
It mainly manifests in two primary aspects. 
First, although Graph Neural Networks (GNNs) have been widely adopted to capture inter-channel relationships~\cite{DING2023527,10.1016/j.neucom.2024.129024}, they still lack an explicit mechanisim to constrain the side effects of over-generalization in the spatial domain.
Second, the perception of spatial anomalies is rarely incorporated into the final anomaly scoring.
For the latter, the pivotal obstacle lies in the absence of a dedicated benchmark to comprehensively evaluate the capability of models in localizing the specific channels where anomalies occur.
Existing MTSAD datasets typically provide only time-wise binary labels, omitting the spatial footprint of faults.
However, in practical AD scenarios, channel-wise anomaly localization holds substantial significance for result interpretability and root-cause attribution.

To bridge these gaps, we propose a novel MTSAD framework based on Prior–Observation adversarial learning of Spatio–Temporal associations (POST). 
To address the learning of spatial dependencies, we restructure the standard Transformer architecture, mitigating the information entanglement caused by positional cues.
On top of that, we propose the Spatial Anomaly Graph Attention (SAGA) module. 
Integrating Graph Structure Learning (GSL), SAGA formulates a regularized adjacency matrix as a structural prior and conducts minimax adversarial learning against data-driven spatial observations.
To explicitly evaluate the capacity of models in learning spatial topologies, we construct a challenging synthetic benchmark based on the Server Machine Dataset (SMD), named SMD+.
Equipped with exact channel-wise anomaly labels, SMD+ imposes strict evaluation criteria for anomaly localization.
In summary, the main contributions of this paper are threefold:
\begin{itemize}
\item We propose POST, a novel MTSAD model that unifies spatial and temporal anomaly modeling under a joint prior–observation adversarial learning framework. 
\item We construct SMD+, a dataset featuring precise channel-wise annotations. The dataset can serve as a generic testbed for anomaly localization in MTS. 
\item Extensive experiments demonstrate that POST consistently outperforms state-of-the-art (SOTA) baselines in standard anomaly detection tasks. Moreover, the optimization of spatial association discrepancy yields superior anomaly localization, significantly enhancing the diagnostic interpretability of final results.
\end{itemize}

\section{Related Work}
\subsection{Multivariate Time Series Anomaly Detection}
The evolution of unsupervised MTSAD has been profoundly shaped by deep representation learning.
Early deep architectures, such as RNNs~\cite{s23249878,10.1145/3219819.3219845,10.1145/3292500.3330672} and AEs~\cite{9064715}, were predominantly employed to capture sequential dependencies. 
As modern industrial systems grow in complexity, variables (sensors) rarely operate in isolation. 
Consequently, spatio-temporal MTSAD methods have emerged to simultaneously model temporal dynamics and spatial inter-sensor topologies.
A prevailing trend is the integration of GNNs with sequence models.
For instance, TopoMAD~\cite{9228885} utilizes GNNs to extract spatial features and LSTMs to capture temporal dependencies. 
Approaches such as STGAT-MAD~\cite{9747274}, MST-GAT~\cite{DING2023527}, and TopoGDN~\cite{10.1145/3627673.3679614} explicitly employ graph attention or multi-modal graph structures to guide signal reconstruction. 
Furthermore, STAT~\cite{FU2024119978} integrates anomaly attention~\cite{xu2022anomaly} with transposed attention modules to directly model temporal dynamics and topological structures.

Despite these advancements, existing studies of sptatial topologies in MTSAD frameworks suffer from two essential limitations. First, \textit{spatio-temporal entanglement}: conventional methods typically adopt the standard Transformer framework with global Positional Encoding (PE) across the input representation. 
In this case, the temporal positional cues inevitably contaminate the spatial graph learning, causing severe information entanglement.
Second, \textit{unconstrained spatial reconstruction}: most existing works rely either on static, statistically derived adjacency matrices or unconstrained dynamic graphs. 
Without structural constraints, these spatial models can easily over-generalize and perfectly reconstruct anomalies as well, leading to degradation in recall performance
In this paper, we theoretically discuss these limitations and propose dedicated mechanisms to construct a more robust spatio-temporal modeling of latent associations. 

\subsection{Learning Spatial Topology}
Techniques for learning complex dependencies among multiple variables can be traced back to early feature recalibration mechanisms in convolutional neural networks.
For instance, channel-attention modules such as SENet~\cite{8578843}, CBAM~\cite{10.1007/978-3-030-01234-2_1}, and ECA-Net~\cite{9156697} utilize squeeze-and-excitation or lightweight convolutions to adaptively re-weight channel responses.
While computationally efficient, these approaches are inherently implicit. 
They adjust individual feature magnitudes without establishing a formal topology among variables. 
In the context of MTSAD, where anomalies often manifest as structural deviations between specific sensors, this lack of relationship modeling is a significant drawback.
Consequently, explicitly capturing these relationships via GNNs~\cite{9046288,DBLP:conf/iclr/KipfW17}, particularly Graph Attention Networks (GATs)~\cite{velickovic2018graph}, has become a prevailing trend in the literature of MTS analysis.
The adjacency matrix in these modules directly represents the prior knowledge of spatial dependencies, which provides the foundation for the comparison between prior and observation.

Despite these advantages, standard GNNs present a fundamental limitation.
The heavily rely on predefined, static adjacency matrices.
In most practical scenarios, an accurate physical prior of the spatial structure is difficult to obtain.
As a compromise, existing MTSAD methods~\cite{9747274,DING2023527,10.1145/3627673.3679614} typically choose to derive the adjacency matrix in a statistical manner from the training data.
The potential risk of this approach is that statistical estimates are not always reliable.
Once fixed, they force the model to adapt under a potentially biased topology.
To resolve this, the GNN community has increasingly adopted the framework of GSL.
LDS~\cite{franceschi2019learning} formulates the joint inference of discrete graph structures and GNN parameters as a bilevel optimization problem. 
Similarly, Pro-GNN~\cite{jin2020graph} learns a robust data-driven topology by explicitly enforcing intrinsic structural properties, such as sparsity and low rank.
Our POST framework integrates this data-driven mechanism into the proposed SAGA module.
However, note that SAGA still diverges significantly from standard GAT formulations.
To natively support prior-observation adversarial learning in the spatial domain, we thoroughly redesign the algorithm, proposing a complete procedure of feedforward inference as well as the alternating optimization between the adjacency matrices and the network parameters.

\section{Preliminaries}
\label{sec:pre}
In this section, we first briefly revisit the multivariate time-series anomaly detection problem. 
Formally, let the multivariate time series be denoted as $\mathcal{W} = \{\boldsymbol\omega_1, \boldsymbol\omega_2, \ldots, \boldsymbol\omega_T\}$, where each observation $\boldsymbol\omega_t \in \mathbb{R}^{D_0}$ corresponds to the measurements of $D_0$ variables (or sensors) at time step $t$. 
Following the common unsupervised setting in MTSAD, we assume that the training set contains only normal samples. The objective is to learn a representation of normal system behavior such that, during inference, the model can assign an anomaly score to each time step and identify abnormal events accordingly.

Specifically, we divide the signal into non-overlapping input segments denoted as $\mathcal{W}_t = \big[\boldsymbol\omega_i\big]_{i=t-N+1}^t$, where $N$ is the window length covering the observation at time step $t$ and its $N-1$ predecessors. 
The model takes $\mathcal{W}_t$ as input and computes the anomaly scores for all time points within the segment. 
Higher scores indicate a greater likelihood of anomalies. 
The final decision is made by thresholding the point-wise anomaly score.

\textbf{Anomaly Transformer}:
AT establishes a typical adversarial learning framework for MTSAD based on the Transformer architecture. 
First, the raw observation sequence is embedded by linear projection as $\boldsymbol{X}=\big[\boldsymbol{x}_i\big]_{i=1}^N$, where each $\boldsymbol{x}_i\in\mathbb{R}^{D}$ denotes the embedded feature, and $D$ is the dimension of feature space.
To incorporate temporal order into the embeddings, it applies Absolute Positional Encoding (APE):
\begin{equation}
\label{eq:pe}
  \boldsymbol{x}_i^1 = f_{ape}(\boldsymbol{x}_i) = \boldsymbol{x}_i + \boldsymbol{p}_i,
\end{equation}
where $\boldsymbol{p}_i\in\mathbb{R}^{D}$ is the sinusoidal encoding vector, defined as
\begin{equation}
  \begin{cases} 
    p_{i,2d}    = \sin(i/10000^{2d/D}) \\ 
    p_{i,2d+1}  = \cos(i/10000^{2d/D}). 
\end{cases}
\label{eq:ape}
\end{equation}
Then AT performs reconstruction-based anomaly detection within the Multi-Head Self Attention (MHSA) framework. 
At the $l$-th layer of the encoder, given encoded input $\boldsymbol{X}^{l}$, query, key, and value representations are calculated as following:
\begin{equation}
  \boldsymbol{Q}^l = \boldsymbol{X}^{l}\boldsymbol{W}_Q^l, \quad
  \boldsymbol{K}^l = \boldsymbol{X}^{l}\boldsymbol{W}_K^l, \quad
  \boldsymbol{V}^l = \boldsymbol{X}^{l}\boldsymbol{W}_V^l, \quad
\end{equation}
where the functions $\boldsymbol{W}_Q^l, \boldsymbol{W}_K^l, \boldsymbol{W}_V^l$ are learnable linear projections. The attention weights $\mathcal{S}^l$ and the reconstruction $\boldsymbol{X}_t^l$ of input is then given by
\begin{equation}
  \label{eq:att}
\mathcal{S}^l = \text{Softmax}(\frac{\boldsymbol{Q}^{l} \boldsymbol{K}^{l\top}}{\sqrt{d_q}}),
\end{equation}
\begin{equation}
  \boldsymbol{X}_t^l = \mathcal{S}^l\boldsymbol{V}^l,
\end{equation}
where $d_{q}$ is the dimension of the query.
To detect anomalies, AT proposed anomaly attention by augmenting the standard attention with a prior–observation association discrepancy. In each layer $l$, a prior distribution $\mathcal{P}^l$ is defined by a rescaled Gaussian kernel as
\begin{equation}
  \label{eq:p_gauss}
  \mathcal{P}^l = \text{Rescale}\Big(\Big[\frac{1}{\sqrt{2\pi}\sigma_i^l}\exp\!\big(-\tfrac{|j-i|^2}{2(\sigma_i^l)^2}\big)\Big]_{i,j=1}^N\Big), 
\end{equation}
where $\sigma_i^l=\boldsymbol{W}_{\sigma}^{l\top}\boldsymbol{x}_i^{l}$ is a learnable kernel bandwidth also derived from the input.
Finally, the association discrepancy between attention-derived observation $\mathcal{S}^l$ and Gaussian-based prior $\mathcal{P}^l$ is defined as
\begin{equation}
  \label{eq:assdis_t}
  \text{AssDis}_t(\mathcal{P},\mathcal{S})= \Big[\frac{1}{L}\sum_l\text{KL}_{\text{sym}}(\mathcal{P}^l_{i:}\,\|\,\mathcal{S}^l_{i:})\Big]_{i=1}^N,
\end{equation}
where $\text{KL}_{\text{sym}}(\mathcal{P}^l_{i:}\,\|\,\mathcal{S}^l_{i:})=\text{KL}(\mathcal{P}^l_{i:}\,\|\,\mathcal{S}^l_{i:})+\text{KL}(\mathcal{S}^l_{i:}\,\|\,\mathcal{P}^l_{i:})$ denotes the symmetric Kullback-Leibler divergence.
After $L$ layers, the final representation $\boldsymbol{X}_t^L$ is projected back to the original data space via a linear layer to obtain the reconstruction $\big[\tilde{\boldsymbol\omega}_i\big]_{i=1}^N$. 

The training of this model alternates between two phases: (i) detaching $\mathcal{S}^l$ and updating $\mathcal{P}^l$ to minimize entries in $\text{AssDis}_t$, making the prior adapt to diverse patterns; (ii) detaching $\mathcal{P}^l$ and updating attention to maximize entires in $\text{AssDis}_t$, preventing trivial self-correlation and enforcing non-local associations. Finally, anomaly score vector $\text{AS}(\mathcal{W}_t)$ is obtained by combining reconstruction error with averaged association discrepancy:
\begin{equation}
  \label{eq:as_at}
  \begin{split}
    \text{AS}(\mathcal{W}_t) &= \text{Softmax}\big(-\text{AssDis}_t(\mathcal{P},\mathcal{S})\big) \\
    &\quad \odot \Big[\|\tilde{\boldsymbol\omega}_i-\boldsymbol\omega_i\|_2^2\Big]_{i=1}^N ,
  \end{split}
\end{equation}
where $\odot$ is the element-wise multiplication and $\|\cdot\|_2$ denotes the Euclidean norm. $\text{AS}(\mathcal{W}_t)\in\mathbb{R}^{N\times1}$ represents the point-wise anomaly criterion of each $\boldsymbol\omega_i$ in original signal.

The design of AT is motivated by the limitation that training relies only on normal samples. If reconstruction is enforced without additional constraints, the attention mechanism may degenerate into trivial self-correlation, where each time step is simply reconstructed from itself. Such behavior allows the model to reconstruct both normal and abnormal inputs, thereby losing discriminative capability. By introducing the prior association $\mathcal{P}$, AT prevents this trivial solution and compels the model to exploit meaningful cross-time associations. Consequently, anomalies can be detected when either the reconstruction error increases or the association discrepancy decreases.
It is worth noting that the capacity of AT does not necessarily increase detection performance: deeper layers may still yield nearly perfect separation between $\mathcal{P}^l$ and $\mathcal{S}^l$ even for anomalies, making the model rely solely on reconstruction error without theoretical guarantees. This observation highlights that anomaly detection benefits more from carefully designed constraints such as $\text{AssDis}$, rather than arbitrarily increasing model complexity.

\section{The Proposed Framework}
\begin{figure*}[!t]
  \centering
  \includegraphics[width=1.0\textwidth]{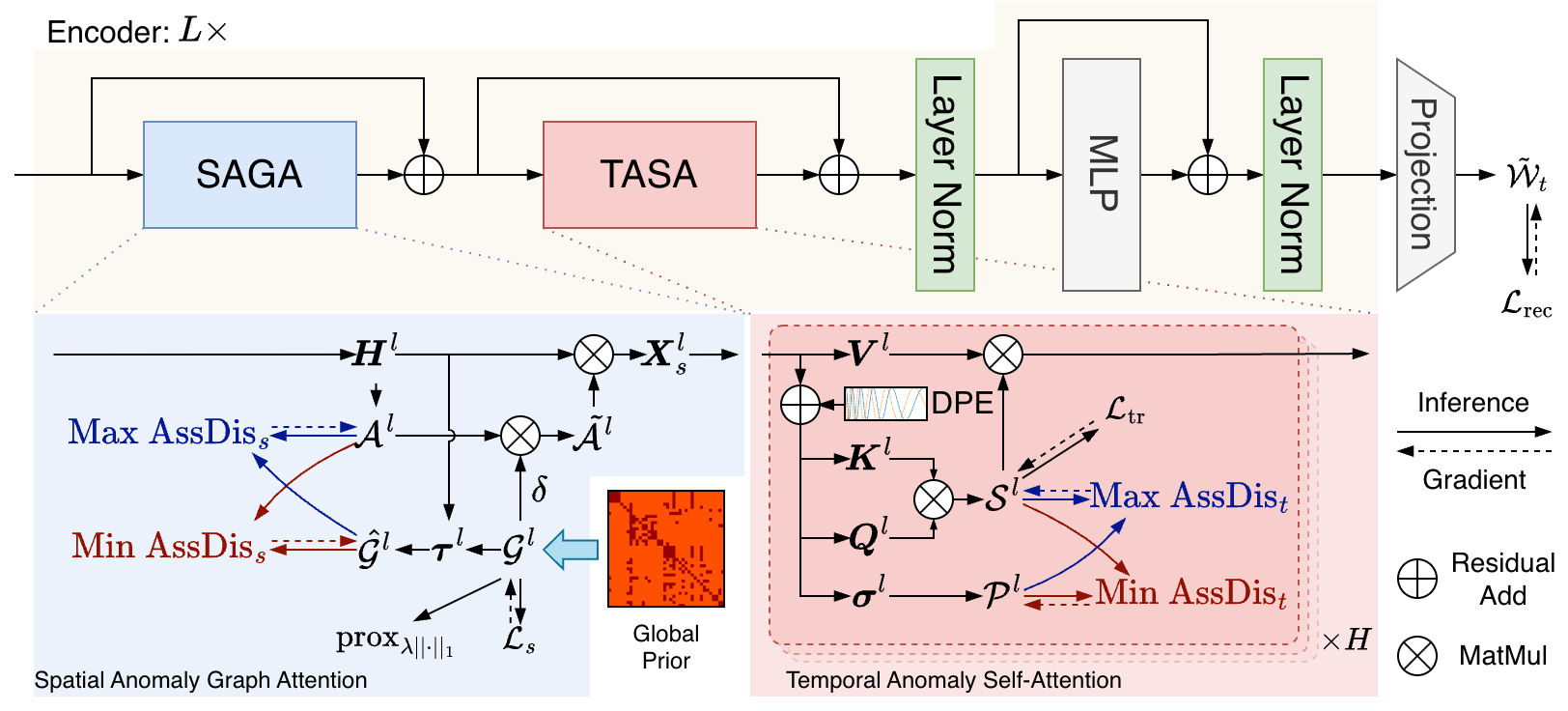}
  \caption{Overall architecture of the proposed POST framework. The model alternates between the Spatial Anomaly Graph Attention (SAGA) and the Temporal Anomaly Self-Attention (TASA) modules, followed by normalization and feed-forward layers. SAGA learns a data-driven adjacency matrix regularized by sparsity and smoothness, and combines it with graph attention to capture spatial dependencies. TASA introduces positional encoding only in the anomaly-attention mechanism and enforces a prior–observation adversarial training scheme to capture temporal discrepancies. The outputs are aggregated through multi-layer processing and projected back to the original space for reconstruction and anomaly scoring.  
}
  \label{fig:framework}
  \end{figure*}

In this section, we present the overall architecture of our proposed framework. 
As illustrated in Figure~\ref{fig:framework}, the framework extends the AT by introducing two key enhancements. 
First, in the temporal dimension, we propose Temporal Anomaly Self-Attention (TASA) with a disentangled positional encoding mechanism and an additional symmetric KL-based regularization term, ensuring that the structure of anomaly attention to be compatible with joint spatio–temporal modeling. 
More importantly, we introduce the Spatial Anomaly Graph Attention (SAGA) module, which performs prior–observation adversarial learning over sensor dependencies and explicitly models spatial associations. 
By combining these two extensions, our proposed POST unifies spatial and temporal anomaly modeling under a joint adversarial learning framework, achieving a more accurate detection of multivariate anomalies.

\subsection{Temporal Anomaly Self-Attention}
As illustrated in the lower-right part of Fig.~\ref{fig:framework}, we propose the Temporal Anomaly Self-Attention (TASA) module by introducing several critical modifications to the standard anomaly attention. 
This module strictly confines the influence of the positional encoding mechanism on signal reconstruction within TASA itself, thereby enabling the overall framework to jointly model both spatial and temporal dependencies. 
The primary challenge addressed here arises from the use of APE, as defined in Eq.~(\ref{eq:pe}). 
The original anomaly attention employs APE to facilitate the adversarial interaction between the series association $\mathcal{S}$ and the prior association $\mathcal{P}$. 
As formulated in Eq.~(\ref{eq:p_gauss}), $\mathcal{P}$ serves as a Gaussian kernel function whose center is strictly tied to the temporal position $i$ of the input $\boldsymbol{x}_i$. 
During the adversarial process, the model naturally leverages positional encodings to perceive the relative distances among inputs, thereby pushing the peak of $\mathcal{S}$ away from that of $\mathcal{P}$. 
This mechanism is crucial for the stability of the adversarial framework, and our experimental observations also confirm this behavior. 

However, from the perspective of signal reconstruction, directly adding positional information to the raw inputs $\boldsymbol{x}_i$ of model actually introduces noise into the representations. 
Particularly for spatial modeling modules, temporal positional encoding merely acts as irrelevant noise, thereby degrading the reconstruction quality. 
To address this issue, we propose the Disentangled Positional Encoding (DPE) mechanism. 
DPE ensures that the effect of positional encoding is confined within the TASA module at each layer, without influencing the rest part of the model.

Formally, we denote the input of the TASA module at layer $l$ as $\boldsymbol{x}_i^{l}$. The feature is augmented with positional encoding as
\begin{equation}
  \label{eq:dpe}
  \hat{\boldsymbol{x}}_i^{l} = \boldsymbol{x}_i^{l} + \boldsymbol{p}_i,
\end{equation}
where the calculation of $\boldsymbol{p}_i$ is consistent with Eq.~(\ref{eq:ape}).
With the position-aware input $\hat{\boldsymbol{X}}^{l}$, the branches in the TASA module can be computed respectively as
\begin{equation}
  \begin{aligned}
    \boldsymbol{Q}^l &= \hat{\boldsymbol{X}}^{l}\boldsymbol{W}_Q^l,\quad
    \boldsymbol{K}^l = \hat{\boldsymbol{X}}^{l}\boldsymbol{W}_K^l,\quad \\
    \boldsymbol{V}^l &= \boldsymbol{X}^{l}\boldsymbol{W}_V^l,\quad
    \boldsymbol{\sigma}^l = \hat{\boldsymbol{X}}^{l}\boldsymbol{W}_{\sigma}^l.
  \end{aligned}
\end{equation}
The remaining computations follow Eq.~(\ref{eq:att})–(\ref{eq:assdis_t}). 
One can see that here DPE is injected only into the $\boldsymbol{Q}$ and $\boldsymbol{K}$ branches to calculate $\mathcal{S}$. 
The final reweighting still acts on the original features, hence clean feature representations will flow into subsequent layers.

DPE is reminiscent of the Rotary Positional Encoding (RoPE) method~\cite{su2023roformerenhancedtransformerrotary}, which also restricts position encoding to the computation of $\mathcal{S}$. 
The main difference is that DPE retains absolute positional information by directly adding $\boldsymbol{p}_i$ as in Eq.~(\ref{eq:ape}), rather than encoding relative positions through rotations. 
This design is motivated by our observation that RoPE is inappropriate in the scenario of MTSAD. 
The main advantage of RoPE lies in directly modeling relative positions in long sequences to ease training, while in MTSAD the sequence length is fixed and relatively short.
Moreover, RoPE retains only relative position information and completely removes absolute positions, which hinders accurate signal reconstruction.
Therefore, we choose to follow Eq.~(\ref{eq:ape}) and add absolute positional encodings to the corresponding branches.

\textbf{Regularization of attention scores}: 
The design of adversarial learning in anomaly attention aims to discover non-local associations between normal signals by optimizing Eq.~(\ref{eq:assdis_t}).
However, we found that model sometimes learns $\mathcal{S}$ distributions with peaks fixed at certain positions, independent of specific input content.
Although these positions are deliberately pushed far away from the peaks of $\mathcal{P}$ to successfully maximize the objective in Eq.~(\ref{eq:assdis_t}), we still treat such identical $\mathcal{S}$ shared by different samples as trivial solution since they are apparently insensitive to whether anomaly exists.
To avoid such trivial solutions, we introduce a regularization term to further constrict the consistence of $\mathcal{S}$.
To capture the overall behavior of each head in the attention module, we aggregate the row-wise attention scores as
\begin{equation}
  \label{eq:agg_s}
  \bar{\mathcal{S}}^{l,b,h} = \frac{1}{N}\sum_{i=1}^{N} \mathcal{S}^{\,l,b,h}_{i:},
\end{equation}
where $b$ and $h$ denote the batch and head indices, respectively. 
We only use the notation $\bar{\mathcal{S}}^{l,b,h}$ in this section to explicitly distinguish between different heads and samples, elsewhere we retain the simpler notation of $\mathcal{S}^l$ with no ambiguity.
Based on the aggregated distribution, we define a symmetric KL divergence–based triplet loss:
\begin{equation}
  \label{eq:reg_tr}
  \begin{aligned}
    \mathcal{L}_{\text{tr}} =\frac{1}{LH}\sum_{h=1}^{H}\sum_{l=1}^{L}\big[m &+ \min_{h'\neq{h}}\text{KL}_{\text{sym}}(\bar{\mathcal{S}}^{l,b,h}\,\|\,\bar{\mathcal{S}}^{l,b,h'})\\
                                         &- \text{KL}_{\text{sym}}(\bar{\mathcal{S}}^{l,b,h}\,\|\,\bar{\mathcal{S}}^{l,b',h})\big]_+
  \end{aligned}
\end{equation}
where $b'\neq{b}$ is a randomly sampled instance within the batch, $m$ is a small positive margin and $[\cdot]_+$ represents the hinge function.
This loss enforces that intra-sample similarity across different heads of the same instance must exceed inter-sample similarity from different instances.

\subsection{Spatial Anomaly Graph Attention}
\label{sec:saga}
To extend the prior-observation adversarial learning into the spatial domain, we propose the Spatial Anomaly Graph Attention (SAGA) module. 
SAGA achieves the signal reconstruction by learning cross-channel associations within a Graph Attention Network (GAT) framework~\cite{velickovic2018graph}, while simultaneously executing an adversarial learning between the structural prior and the data-driven observation.
Unlike temporal dependencies where a Gaussian prior can naturally capture local Markovian properties, relationships between different sensors generally lack an inherent prior.
To address this, SAGA employs a layer-wise learnable adjacency matrix $\mathcal{G}^l \in \mathbb{R}^{D_0 \times D_0}$ to represent the prior structural knowledge regarding sensor associations.
Furthermore, we introduce a learnable temperature parameter $\boldsymbol{\tau}^l$ to control the concentration of the prior distribution.
Analogous to the kernel bandwidth $\boldsymbol{\sigma}^l$ in the TASA module, $\boldsymbol{\tau}^l$ allows for the dynamic adjustment of the learned prior graph, effectively scaling its sharpness.

Given the input of the $l$-th layer $\boldsymbol{X}^{l}\in\mathbb{R}^{N\times{D}}$, SAGA first projects it into the raw channel space of dimension $D_0$:
\begin{equation}
  \label{eq:projection}
\boldsymbol{H}^l=(\boldsymbol{X}^{l}\boldsymbol{W}_H^l)^\top,
\end{equation}
where $\boldsymbol{W}_H^l \in \mathbb{R}^{D \times D_0}$ is a learnable projection matrix.
Then, the distribution of observation attention $\mathcal{A}^l$ is computed as
\begin{equation}
\label{eq:saga_a}
\mathcal{A}^l_{ij} = \frac{\exp\big(\text{LeakyReLU}\big([\boldsymbol{H}_{i:}^l | \boldsymbol{H}_{j:}^l]\boldsymbol{\theta}^l\big)\big)}{\sum_{n=1}^{D_0}\exp\big(\text{LeakyReLU}\big([\boldsymbol{H}_{i:}^l | \boldsymbol{H}_{n:}^l]\boldsymbol{\theta}^l\big)\big)},
\end{equation}
where $\boldsymbol{\theta}^l\in\mathbb{R}^{2N}$ is learnable parameter vector, $[\cdot|\cdot]$ represents the concatenation operator.
An activation function of leaky ReLU is adopted here by following the generic design of standard GAT.
Then attention $\mathcal{A}^l$ is combined with the prior to obtain the posterior distribution of attention as
\begin{equation}
  \label{eq:saga_a_posterior}
  \tilde{\mathcal{A}}_{i:}^l=\tilde{\mathcal{G}}_{i:}^l\odot\mathcal{A}_{i:}^l/Z_i^l,
\end{equation}
where $Z_i^l$ is the partition factor, $\tilde{\mathcal{G}}^l=\delta(\mathcal{G}^l)$ is the Bernoulli-form distribution of prior $\mathcal{G}^l$, and $\delta$ denotes the Sigmoid function.
The posterior $\tilde{\mathcal{A}}^l$ is used to reweight the features $\boldsymbol{H}^l$ and achieve the reconstruction $\boldsymbol{X}_s^{l}$ of input.
Combining all the above steps, the whole procedure of SAGA is shown in Algorithm~\ref{alg:saga}.
\begin{algorithm}[htbp]
  \caption{Spatial Anomaly Graph Attention.}
  \label{alg:saga}
  \begin{algorithmic}[1]
  \setlength{\itemsep}{4pt}          
  \REQUIRE Input $\boldsymbol{X}^{l}$, learnable adjacency matrix $\mathcal{G}^l$.
  \ENSURE Updated reconstruction $\boldsymbol{X}_s^{l}$.
  \STATE $\boldsymbol{H}^l=(\boldsymbol{X}^{l}\boldsymbol{W}_H^l)^\top,\ 
  \boldsymbol{W}_H^l\in\mathbb{R}^{D\times{D_0}}$ \hfill \{Projection\}
  \STATE $E_{ij}^l = \text{LeakyReLU}\big([\boldsymbol{H}_{i:}^l | \boldsymbol{H}_{j:}^l]\boldsymbol{\theta}^l\big)$ \hfill \{Spatial association\}
  \STATE $\mathcal{A}^l = \text{Softmax}(\boldsymbol{E}^l)$ \hfill \{Attention\}
  \STATE $\tilde{\mathcal{G}}^l= \delta(\mathcal{G}^l)$ \hfill \{Prior\}
  \STATE $\tilde{\mathcal{A}}_{ij}^l = \tilde{\mathcal{G}}^l_{ij}\mathcal{A}_{ij}^l/\sum_{n=1}^{D_0}\tilde{\mathcal{G}}^l_{in}\mathcal{A}_{in}^l$ \hfill \{Posterior $\tilde{\mathcal{A}}^l$\}
  \STATE $\boldsymbol{X}_s^{l}=(\tilde{\mathcal{A}}^l\boldsymbol{H}^l)^\top \boldsymbol{W}_S^l$,
  $\boldsymbol{W}_S^l\in\mathbb{R}^{D_0\times{D}}$ \hfill \{Reconstruction\}
  \end{algorithmic}
  \end{algorithm}

Besides learnable adjacency prior, reweighting the features with the posterior $\tilde{\mathcal{A}}^l$ in Eq.~(\ref{eq:saga_a_posterior}) is a main difference between SAGA and standard GAT.
It brings two main benefits. First, from an inference perspective, signal reconstruction here is not only driven by local observations but also aligned with the statistical priors exhibited on the training set.
Second, it allows $\mathcal{G}^l$ to be effectively learned and dynamically capture the characteristics of training samples, rather than being restricted to fixed values.

\textbf{Adversarial learning in SAGA}: 
To implement the prior-observation adversarial learning in the spatial domain, we explicitly optimize the spatial association discrepancy $\text{AssDis}_s$ as:
\begin{equation}
  \text{AssDis}_s(\hat{\mathcal{G}},\mathcal{A})= \Big[\frac{1}{L}\sum_l\text{KL}_{\text{sym}}(\hat{\mathcal{G}}^l_{i:}\,\|\,\mathcal{A}^l_{i:})\Big]_{i=1}^{D_0},
  \end{equation}
where $\hat{\mathcal{G}}_{i:}^l=\text{Softmax}(\mathcal{G}_{i:}^l/\tau_i^l)$, and $\tau_i^l \in (0, 1)$ is the temperature derived from $\boldsymbol{H}_{i:}^l$ via learnable projection $\boldsymbol{W}_{\tau}^l$.
This objective involves the joint optimization of the adjacency graph $\mathcal{G}^l$, the temperature $\boldsymbol{\tau}^l$, and the observation attention $\mathcal{A}^l$. 
To manage this complex parameter space, we integrate the minimax adversarial optimization with the standard graph structure learning framework~\cite{franceschi2019learning,jin2020graph}, jointly learning $\mathcal{G}^l$ and other model parameters using different optimizers.
Specifically, during the minimization phase, we optimize $\hat{\mathcal{G}}^l$ by updating $\mathcal{G}^l$ and $\boldsymbol{\tau}^l$ with a detached $\mathcal{A}^l$. This ensures that, from the prior perspective, the distribution of the most correlated sensors aligns with the likelihoods derived from the input samples. 
Conversely, during the maximization phase, we optimize $\mathcal{A}^l$ with a detached $\hat{\mathcal{G}}^l$. This encourages the model to capture broader associations rather than collapsing into trivial self-correlations.

During inference, the synergy between static $\mathcal{G}^l$ and the input-driven $\boldsymbol{\tau}^l$ provides the flexibility to discover non-trivial dependencies that are solely exhibited by normal training samples.
Under normal conditions, $\mathcal{G}^l$ robustly reflects the statistical correlations across the training set, and the model maintains a smooth temperature $\boldsymbol{\tau}^l\to 1$ to align $\hat{\mathcal{G}}^l$ with the observed attention $\mathcal{A}^l$. 
Conversely, when an anomaly occurs, the input features deviate from the normal distribution. 
Although $\mathcal{G}^l$ remains stable, the perturbed $\boldsymbol{\tau}^l$ tends to compel the scaled prior $\hat{\mathcal{G}}^l$ to collapse toward a sharp peak on the self-loop. 
Meanwhile, the observation attention $\mathcal{A}^l$ loses reliable statistical significance and collapses toward a similar trivial pattern as well.
As a consequence, the discrepancy $\text{AssDis}_s$ decreases and serves as a discriminative criterion for structural deviations in the spatial domain.
It is also worth noting that we deliberately omit any form of positional encoding from SAGA. 
Differing from temporal signals, the relative index positions between sensor channels lack meaningful distance semantics. 
Therefore, introducing spatial positional encoding provides no benefit for optimizing $\text{AssDis}_s$.

\textbf{Structural regularization on adjacency graphs}:
Inspired by the graph structure learning, we also introduce constraints to $\mathcal{G}^l$ to emphasize its role as a topological prior rather than solely data-driven parameters.
In detail, we adopt two regularization terms.
The first is a smoothness constraint across all layers:
\begin{equation}
  \mathcal{L}_s=\frac{1}{L}\sum_l\text{tr}(\mathcal{W}_t^{\top}\boldsymbol{\Delta}^l\mathcal{W}_t),
\end{equation}
where $\mathcal{W}_t$ represents the original input series, and 
\begin{equation}
  \boldsymbol{\Delta}^l = (\boldsymbol{D}^l)^{-1/2}(\boldsymbol{D}^l - \tilde{\mathcal{G}}^l)(\boldsymbol{D}^l)^{-1/2}
\end{equation}
is the normalized Laplacian matrix for the $l$-th layer, and $\boldsymbol{D}^l$ is the degree matrix of $\tilde{\mathcal{G}}^l$.
The effect of $\mathcal{L}_s$ is to compress the weights in $\tilde{\mathcal{G}}^l$ towards the diagonal, thereby ensuring that self-loops dominate the prior graph structure. 

The second regularization term imposes an $\ell_1$ sparsity constraint on $\tilde{\mathcal{G}}^l$.
Since the sigmoid function guarantees $\delta(\cdot)\in(0,1)$, this regularization is differentiable and can be directly attached to the overall objective.
Nevertheless, in our implementation, we adopt a hybrid strategy inspired by the prevailing Forward-Backward splitting framework~\cite{Combettes2011} in sparse optimization. 
Specifically, we alternate between a gradient descent step and a proximal step applied directly to $\mathcal{G}^l$ instead of $\tilde{\mathcal{G}}^l$.
Given the intermediate update from the gradient descent step for the overall loss function, the proximal step is to solve the following problem:
\begin{equation}
  \label{eq:prox}
  \text{prox}_{\lambda\|\delta(\cdot)\|_1}(\mathcal{G}^l) = \operatorname*{arg\,min}_{\mathcal{Z}\in \mathbb{R}^{D_0\times{D_0}}} \frac{1}{2}\|\mathcal{Z}-\mathcal{G}^l\|_F^2 + \lambda\|\delta(\mathcal{Z})\|_1,
\end{equation}
where $\|\cdot\|_F$ denotes the Frobenius norm.
Unlike the standard $\ell_1$ proximal operator, Eq.~(\ref{eq:prox}) lacks a closed-form solution due to the non-linear composite term. 
However, it can be proven that the problem is strictly convex for sufficiently small $\lambda$. 
The detailed derivation can be found in the supplementary material.
Consequently, we choose to optimize it via fixed-point iteration as
\begin{equation}
  \label{eq:prox_iter}
  \mathcal{Z}_{ij}^{(k+1)} = \mathcal{G}_{ij}^l-\lambda\delta(\mathcal{Z}_{ij}^{(k)})\left(1-\delta(\mathcal{Z}_{ij}^{(k)})\right).
\end{equation}
The iteration is executed element-wise, analogously to the standard $\ell_1$ proximal operator.
This operator provides consistent gradient updates to the aforementioned direct incorporation of the $\ell_1$ penalty.
The difference is the fixed-point iteration can be performed for a few steps and finally converges to a more robust solution.
Our experiments in Section~\ref{sec:exp_l1} indicate that this proximal scheme provides incremental but stable improvements over the plain $\ell_1$ regularization.
With the above regularization terms, the optimization of $\mathcal{G}^l$ is explicitly decoupled from the observation attention $\mathcal{A}$, thereby ensuring that the adversarial learning within SAGA remains robust and avoids trivial solutions.

\subsection{Model Training}
\label{sec:training}
There are two main groups of parameters need to be optimized in our proposed model: the layer-specific adjacency matrices $\mathcal{G}^l$s, and the learnable network parameters $\boldsymbol{W}$ which include $\boldsymbol{W}_{\sigma}$ and $\boldsymbol{W}_{\tau}$ for generating the dynamic temporal bandwidths and spatial temperatures, respectively.
We adopt an alternating optimization scheme~\cite{jin2020graph,franceschi2019learning} to iteratively optimize the aforementioned loss functions and update these parameters.
The complete training procedure is summarized in Algorithm~\ref{alg:train}.
\begin{algorithm}[htbp]
  \caption{Training procedure of the proposed model.}
  \label{alg:train}
  \begin{algorithmic}[1]
  \REQUIRE Training set $\mathcal{W}$, hyper-parameters $\alpha, \beta, \gamma, \xi, \lambda$
  \ENSURE Learned parameters $\boldsymbol{W},\boldsymbol{W}_{\sigma},\boldsymbol{W}_{\tau}$, and adjacency matrices $\mathcal{G}^l$s
  \STATE Randomly initialize $\boldsymbol{W},\boldsymbol{W}_{\sigma},\boldsymbol{W}_{\tau}$
  \STATE Initialize $\mathcal{G}^l\leftarrow k\text{NN}(\mathcal{W})$
  \WHILE {epoch $<$ max\_epochs}
  \STATE Update $\mathcal{G}^l$ by optimizing: 
  $$\mathcal{L}_{\text{rec}}+\beta\|\text{AssDis}_s(\hat{\mathcal{G}},\mathcal{A}_{\text{detach}})\|_1+\gamma\mathcal{L}_s$$
  \STATE Apply proximal step: 
  $$\mathcal{G}^l\leftarrow\text{prox}_{\lambda\|\delta(\cdot)\|_1}(\mathcal{G}^l)$$
  \FOR {$i=1$ to $5$}
  \STATE Update $\boldsymbol{W}$ by optimizing:
  $$\begin{aligned}
    \mathcal{L}_{\text{rec}}&-\alpha\|\text{AssDis}_t(\mathcal{P}_{\text{detach}},\mathcal{S})\|_1\\
    &-\beta\|\text{AssDis}_s(\hat{\mathcal{G}}_{\text{detach}},\mathcal{A})\|_1+\xi\mathcal{L}_{\text{tr}}
  \end{aligned}$$
  \STATE Update $\boldsymbol{W},\boldsymbol{W}_{\sigma},\boldsymbol{W}_{\tau}$ by optimizing:
  $$\begin{aligned}
    \mathcal{L}_\text{rec}&+\alpha\|\text{AssDis}_t(\mathcal{P},\mathcal{S}_{\text{detach}})\|_1\\
    &+\beta\|\text{AssDis}_s(\hat{\mathcal{G}},\mathcal{A}_{\text{detach}})\|_1
  \end{aligned}$$
  \ENDFOR
  \ENDWHILE
  \end{algorithmic}
  \end{algorithm} 

First, a reconstruction loss $\mathcal{L}_{\text{rec}}$ is formulated as
\begin{equation}
  \label{eq:rec_loss}
  \mathcal{L}_{\text{rec}}=\|\tilde{\mathcal{W}}_t-\mathcal{W}_t\|_F^2,
\end{equation}
where $\tilde{\mathcal{W}}_t$ denotes the final reconstruction output at current time step.
This term serves as the primary data-fidelity objective, which is balanced against the spatial and temporal adversarial learning objectives.
At the beginning of training, we initialize every $\mathcal{G}^l$ at the $l$-th layer with a shared channel-wise $k$-Nearest Neighbor ($k$NN) graph constructed from the training set.
During each training epoch, the general network parameters $\boldsymbol{W}$ are optimized in an inner loop of $5$ iterations.
The adversarial minimax process for $\text{AssDis}_t$ is performed entirely within this inner loop as described in Section~\ref{sec:pre}.

In contrast, the adversarial process for $\text{AssDis}_s$, detailed in Section~\ref{sec:saga}, is decoupled across both the inner and outer loops. 
Specifically, the spatial adjacency prior $\mathcal{G}^l$ is updated in the outer loop, subject to the Laplacian smoothness constraint $\mathcal{L}_s$ and the $\ell_1$ sparsity constraint. 
Meanwhile, the underlying projection weights $\boldsymbol{W}_{\tau}$ are optimized within the inner loop alongside $\boldsymbol{W}_{\sigma}$. 

\subsection{Spatial-Temporal Anomaly Criterion}
\label{sec:scoring}
In the inference phase, we attempt to integrate the learned association discrepancy with the reconstruction error to derive the final anomaly criterion. 
For most existing MTSAD benchmarks, they only provide timestamp-level labels, lacking channel-wise anomaly annotations. 
For these datasets, we follow the common practice defined in Eq.~(\ref{eq:as_at}), solely utilizing the temporal discrepancy $\text{AssDis}_t$ in the final anomaly score. 
In such cases, our proposed spatial association discrepancy $\text{AssDis}_s$ mainly functions as a crucial structural regularizer. 
The adversarial learning within $\text{AssDis}_s$ compels the model to capture robust inter-sensor correlations of normal samples, thereby implicitly enhancing both the reconstruction fidelity and the discriminative capacity of the temporal criterion.

However, we argue that the spatial association discrepancy can actually explicitly improve the final performance of anomaly detection, especially in tasks where accurate localization of anomaly channels matters.
To rigorously evaluate the spatial localization capabilities of the proposed POST framework, we construct a controlled synthetic dataset with strict channel-wise labels, termed SMD+, based on the real-world Server Machine Dataset (SMD). 
Based on the dataset, we propose our complete spatial-temporal association-based anomaly criterion $\text{AS}_{ts}(\mathcal{W}_t)$ as follows:
\begin{align}
  \label{eq:asts}
  \text{AS}_{ts}(\mathcal{W}_t) &= \text{Softmax}\big(-\text{AssDis}_t\big) \otimes \text{Sigmoid}\big(-\text{AssDis}_s\big) \notag \\
  &\quad \odot \Big[(\tilde{\omega}_{ij}-\omega_{ij})^2\Big]_{i,j=1}^{N,D_0},
\end{align}
where $\otimes$ denotes the outer product, which broadcasts the temporal and spatial probabilities to align with the dimensions of the reconstruction error, finally yielding the joint $N \times D_0$ anomaly score matrix.

Differing from the temporal dimension, where anomalies are highlighted by a competitive normalization across time steps, here we adopt the Sigmoid function for the spatial discrepancy $\text{AssDis}_s$. 
We observe that evaluating the anomaly score of each channel independently prevents the masking effect of collective anomalies. 
A more detailed comparison of different activation functions is provided in Section~\ref{sec:exp_atst}.
Furthermore, in practical implementation, we record the running statistics of $\text{AssDis}_s$ derived from the training set and perform Z-score centralization on the raw spatial discrepancies before applying the Sigmoid activation.
Finally, by thresholding this fused score matrix, POST can precisely localize anomalies in both temporal and spatial dimensions.

\section{Experimental Results}
In this section, we systematically compare the overall performance of our proposed POST model with SOTA MTSAD methods across 6 benchmarks to demonstrate its superiority. 
We further conduct extensive ablation studies and hyper-parameter sensitivity analyses to investigate how different spatial and temporal components influence the performance of POST. 
Finally, we provide intuitive case studies and visualizations to gain deeper insights into the underlying mechanisms of our model.

\subsection{Datasets}
\label{sec:datasets}
To comprehensively evaluate the proposed POST framework, we adopt five prevailing public benchmarks alongside a specifically constructed dataset, covering diverse real-world scenarios.
\textit{Secure Water Treatment} (SWaT)~\cite{7469060} is derived from a real water treatment testbed managed by the Public Utilities Board of Singapore. 
It captures 51 sensor dimensions during seven days of normal continuous operations and four days of anomalous scenarios. 
Both provided by NASA~\cite{10.1145/3219819.3219845}, the \textit{Mars Science Laboratory} (MSL) dataset contains 55-dimensional sensor readings collected from the Mars rover, while the \textit{Soil Moisture Active Passive} (SMAP) dataset features 25 dimensions encompassing telemetry information and soil samples gathered by spacecraft. 
\textit{Pooled Server Metrics} (PSM)~\cite{10.1145/3447548.3467174} monitors 25 key metrics (e.g., CPU and memory usage) gathered internally from distributed application server nodes at eBay. 
Similarly, the \textit{Server Machine Dataset} (SMD)~\cite{10.1145/3292500.3330672} records 38-dimensional performance metrics over five weeks from massive compute clusters at a major Internet company, encompassing crucial indicators such as CPU utilization, memory consumption, disk I/O, and network traffic. 
The detailed statistical characteristics of these datasets are summarized in Table~\ref{tab:datasets}.
\begin{table}[htbp]
  \centering
  \caption{Statistical details of MTSAD benchmark datasets. For SMD+, the metrics denote time-wise/channel-wise anomaly ratios.}
  \label{tab:datasets}
  \resizebox{\columnwidth}{!}{
    \begin{tabular}{lcccc}
    \toprule
    \textbf{Dataset} & \textbf{Train Size} & \textbf{Test Size} & \textbf{Dimensions} & \textbf{Anomalies (\%)} \\
    \midrule
    SWaT~\cite{7469060}   & 496,800   & 449,919   & 51 & 11.98 \\
    MSL~\cite{10.1145/3219819.3219845}    & 58,317    & 73,729    & 55 & 10.72 \\
    SMAP~\cite{10.1145/3219819.3219845}   & 135,183   & 427,617   & 25 & 13.13 \\
    PSM~\cite{10.1145/3447548.3467174}    & 132,481   & 87,841    & 25 & 27.76 \\
    SMD~\cite{10.1145/3292500.3330672}    & 708,405   & 708,420   & 38 & 4.16 \\
    SMD+   & 708,405   & 609,471   & 38 & 4.35 / 0.18\\
    \bottomrule
    \end{tabular}
  }
\end{table}

\textbf{SMD+ Benchmark}: 
While the aforementioned five benchmarks serve as authoritative testbeds for standard time-wise MTSAD evaluation, the core motivation of our proposed POST framework is to enable comprehensive anomaly localization across both temporal and spatial dimensions. 
To systematically evaluate this spatial anomaly detection capability, we introduce SMD+, a synthetic benchmark built upon the test set of SMD dataset. 
The construction of SMD+ is summarized as follows.
First, to explicitly circumvent the undocumented splicing boundaries inherent in the original data, we extract strictly contiguous, purely normal segments to serve as pristine canvases. 
Second, to simulate diverse realistic failures, we confine the anomaly injection windows exclusively to the tail ends of these segments and define a heterogeneous mixture of transient point and sustained pattern anomalies. 
Finally, following advanced injection protocols~\cite{BENCHMARKS2021_ec5decca}, we implement an additive perturbation mechanism to ensure that the anomalies are deeply convoluted with the inherent system dynamics. 
Detailed descriptions and generation protocols for SMD+ are elaborated in the supplementary material.
Note that SMD+ exclusively shares the uncontaminated training set of the original SMD, enabling a strictly controlled, zero-retraining evaluation.

To emulate the ``needle-in-a-haystack'' nature of early-stage industrial failures, we confine the perturbations to a highly sparse subset of sensor channels (e.g., 1 to 3 out of 38). 
As reflected in Table~\ref{tab:datasets}, while the time-wise anomaly ratio of SMD+ approximately aligns with typical fault occurrences in the original SMD, its channel-wise anomaly ratio is exceptionally low at 0.18\%. 
This numerical discrepancy highlights our strict spatial sparsity constraint, quantifying the difficulty of the benchmark. 
Endowed with this rigorous channel-wise ground truth, SMD+ provides an ideal environment to assess the
model capabilities to localize anomalous channels, as well as the interpretability of their diagnostic outcomes.

\subsection{Implementation Details}
\label{sec:impl}
We stack three layers for both TASA and SAGA modules ($L=3$) to form the Transformer architecture of the POST model. 
The hidden state dimension is set to $D=512$, and the multi-head self-attention mechanism utilizes eight parallel heads ($H=8$). 
For data preprocessing, the time series are segmented using a non-overlapping sliding window, following well-established protocols~\cite{xu2022anomaly,10.5555/3495724.3496816}. 
To accommodate the specific temporal dynamics of different systems, the window size $N$ is set to 200 for the PSM dataset, 150 for SWaT, and 100 for the remaining datasets.
The model is trained for 10 epochs with early stopping using the Adam optimizer with an initial learning rate of $10^{-5}$ and a batch size of 64.
For hyperparameters in Algorithm~\ref{alg:train}, we empirically adopt a generalized setting as $\alpha=0.8,\beta=0.02,\gamma=0.002,\xi=1.0,\lambda=0.7$ via comprehensive grid search.
The detailed parameter sensitivity analysis is deferred to Section~\ref{sec:hyp_param}.

For evaluation, we partition each dataset into an 80\% training set and a 20\% validation set.
The anomaly threshold is dynamically calibrated by statistically analyzing the anomaly scores on the validation set such that a pre-defined proportion $r\%$ of the data is classified as anomalous. We set $r=0.5$ for the SMD and SMD+ datasets and $r=1.0$ for all other datasets~\cite{xu2022anomaly}.
Furthermore, we adopt the standard point-adjustment strategy widely utilized in the MTSAD literature~\cite{10.1145/3178876.3185996,xu2022anomaly,10.5555/3495724.3496816}. 
Under this protocol, if any single time step within a continuous anomalous segment is successfully detected by the model, the entire segment is considered correctly identified. 
The rationale behind this strategy is that a single triggered alert within an anomaly window is typically sufficient to prompt human operators to investigate the entire continuous incident in real-world industrial scenarios. 
For channel-wise spatial evaluation on the SMD+ dataset, this adjustment strategy is repeatedly applied at each channel independently. 
To quantify the detection performance, we employ three standard evaluation metrics: Precision (P), Recall (R), and the harmonic mean, F1-score (F1)~\cite{9618824}.
\begin{table*}[!t]
  \caption{Overall anomaly detection performance comparison on five benchmark datasets. P, R, and F1 denote Precision, Recall, and F1-score (in \%), respectively. The ``Avg'' column represents the average F1-score across all five evaluated datasets. The best results are highlighted in bold, and the second-best results are underlined.
  }
  \label{tab:main_results}
    \centering  
    \resizebox{\linewidth}{!}{  
    \renewcommand{\arraystretch}{1.5}
  \begin{tabular}{l  ccc  ccc  ccc  ccc  ccc  c}                        
  \toprule                                    
  \multicolumn{1}{c}{\multirow{3}*{Method}} & \multicolumn{3}{c}{SMD} & \multicolumn{3}{c}{MSL} & \multicolumn{3}{c}{SMAP} & \multicolumn{3}{c}{SWaT} & \multicolumn{3}{c}{PSM} & \multicolumn{1}{c}{Avg} \cr     
  \cmidrule{2-16}             
   & P & R & F1 & P & R & F1 & P & R & F1 & P & R & F1 & P & R & F1 & F1 \\
  \hline
  VAR~\cite{Anderson1976TimeSeries2E} & 78.35 & 70.26 & 74.08 & 74.68 & 81.42 & 77.90 & 81.38 & 53.88 & 64.83 & 81.59 & 60.29 & 69.34 & 90.71 & 83.82 & 87.13 & 74.66 \\
  LOF~\cite{10.1145/342009.335388} & 56.34 & 39.86 & 46.68 & 47.72 & 85.25 & 61.18 & 58.93 & 56.33 & 57.60 & 72.15 & 65.43 & 68.62 & 57.89 & 90.49 & 0.61 & 46.94 \\
  OCSVM~\cite{Tax2004} & 44.34 & 76.72 & 56.19 & 59.78 & 86.87 & 70.82 & 53.85 & 59.07 & 56.34 & 45.39 & 49.22 & 47.23 & 62.75 & 80.89 & 70.67 & 60.25 \\
  MMPCACD~\cite{7859276} & 71.20 & 79.28 & 75.02 & 81.42 & 61.31 & 69.95 & 88.61 & 75.84 & 81.73 & 82.52 & 68.29 & 74.73 & 76.26 & 78.35 & 77.29 & 75.74 \\
  Deep-SVDD~\cite{pmlr-v80-ruff18a} & 78.54 & 79.67 & 79.10 & 91.92 & 76.63 & 83.58 & 89.93 & 56.02 & 69.04 & 80.42 & 84.45 & 82.39 & 95.41 & 86.49 & 90.73 & 80.97 \\
  DAGMM~\cite{zong2018deep} & 67.30 & 49.89 & 57.30 & 89.60 & 63.93 & 74.62 & 86.45 & 56.73 & 68.51 & 89.92 & 57.84 & 70.40 & 93.49 & 70.03 & 80.08 & 70.18 \\
  LSTM-VAE~\cite{8279425}  & 75.76 & 90.08 & 82.30 & 85.49 & 79.94 & 82.62 & 92.20 & 67.75 & 78.10 & 76.00 & 89.50 & 82.20 & 73.62 & 89.92 & 80.96 & 81.24\\
  LSTM~\cite{10.1145/3219819.3219845} & 78.55 & 85.28 & 81.78 & 85.45 & 82.50 & 83.95 & 89.41 & 78.13 & 83.39 & 86.15 & 83.27 & 84.69 & 76.93 & 89.64 & 82.80 & 83.32\\
  CL-MPPCA~\cite{10.1145/3292500.3330776} & 82.36 & 76.07 & 79.09 & 73.71 & 88.54 & 80.44 & 86.13 & 63.16 & 72.88 & 76.78 & 81.50 & 79.07 & 56.02 & \textbf{99.93} & 71.80 & 76.66\\
  OmniAnomaly~\cite{10.1145/3292500.3330672} & 83.68 & 86.82 & 85.22 & 89.02 & 86.37 & 87.67 & 92.49 & 81.99 & 86.92 & 81.42 & 84.30 & 82.83 & 88.39 & 74.46 & 80.83 & 84.69\\
  BeatGAN~\cite{10.5555/3367471.3367658} & 72.90 & 84.09 & 78.10 & 89.75 & 85.42 & 87.53 & 92.38 & 55.85 & 69.61 & 64.01 & 87.46 & 73.92 & 90.30 & 93.84 & 92.04 & 80.24 \\
  ITAD~\cite{10.1145/3340531.3412716} & 86.22 & 73.71 & 79.48 & 69.44 & 84.09 & 76.07 & 82.42 & 66.89 & 73.85 & 63.13 & 52.08 & 57.08 & 72.80 & 64.02 & 68.13 & 70.92\\
  THOC~\cite{NEURIPS2020_97e401a0} & 79.76 & 90.95 & 84.99 & 88.45 & 90.97 & 89.69 & 92.06 & 89.34 & 90.68 & 83.94 & 86.36 & 85.13 & 88.14 & 90.99 & 89.54 & 88.01\\
  InterFusion~\cite{10.1145/3447548.3467075} & 87.02 & 85.43 & 86.22 & 81.28 & 92.70 & 86.62 & 89.77 & 88.52 & 89.14 & 80.59 & 85.58 & 83.01 & 83.61 & 83.45 & 83.52 & 85.70\\
  TranAD~\cite{10.14778/3514061.3514067}  & 74.30 & 81.65 & 77.80 & 90.72 & 94.73 & 92.68 & 93.12 & 71.33 & 80.78 & 94.23 & 94.36 & 94.29 & 97.44 & 98.19 & 97.92 & 88.69 \\
  AT~\cite{xu2022anomaly}  & 89.40 & 95.45 & 92.33 & 92.09 & 95.15 & 93.59 & 94.13 & \underline{99.40} & \underline{96.69} & 91.55 & 96.73 & 94.07 & 96.91 & 98.90 & 97.89 & \underline{94.91} \\
  MST-GAT~\cite{DING2023527} & - & - & - & \textbf{95.06} & 89.10 & 91.98 & 91.26 & 89.83 & 90.54 & 89.56 & 92.13 & 90.82 & - & - & - & - \\
  ImDiffusion~\cite{chen2023imdiffusion} & \underline{95.20} & 95.09 & \underline{94.88} & 89.30 & 86.38 & 87.79 & 87.71 & 96.18 & 91.75 & 89.88 & 84.65 & 87.09 & 98.11 & 97.53 & 97.81 & 91.86 \\
  STAT~\cite{FU2024119978} & 74.23 & 76.62 & 75.41 & 69.97 & 74.13 & 71.99 & 70.11 & 73.24 & 71.64 & 84.65 & 81.23 & 82.90 & 82.10 & 79.36 & 80.70 & 76.53 \\  
  TopoGDN~\cite{10.1145/3627673.3679614}  & \textbf{97.57} & 83.39 & 89.92 & 82.37 & \textbf{99.67} & 90.23 & - & - & - & 87.93 & 71.91 & 79.11 & - & - & - & - \\
  LGAT~\cite{10.1016/j.neucom.2024.129024} & 88.27 & 85.91 & 87.08 & 91.38 & 90.13 & 90.86 & \underline{94.61} & 92.03 & 93.30 & 93.28 & \underline{99.91} & 96.48 & 98.41 & 98.25 & \underline{98.41} & 93.23 \\
  TSAD~\cite{10.1016/j.neucom.2025.130611}  & - & - & - & 92.21 & 97.76 & \underline{94.91} & - & - & - & 94.24 & 99.30 & \underline{96.70} & 97.53 & 98.43 & 97.98 & - \\
  CSCAD~\cite{10.1016/j.ipm.2025.104315} & 86.21 & \textbf{98.65} & 92.05 & 94.23 & 89.91 & 92.01 & 93.96 & 98.11 & 95.99 & \underline{95.02} & 93.62 & 94.31 & \underline{98.54} & 96.16 & 97.34 & 94.34 \\  
  \midrule
  POST (Ours) & 94.48 & \underline{97.23} & \textbf{95.83} & \underline{94.96} & \underline{98.53} & \textbf{96.71} & \textbf{96.87} & \textbf{99.44} & \textbf{98.14} & \textbf{96.08} & \textbf{100.00} & \textbf{98.00} & \textbf{98.66} & \underline{99.21} & \textbf{98.94} & \textbf{97.52} \\
  \bottomrule
  \end{tabular} 
      }
  \end{table*}

\subsection{Main Comparison on Standard Benchmarks}
To comprehensively evaluate the standard time-wise anomaly detection capabilities of the proposed POST framework, we conduct extensive comparisons on the five public benchmarks against existing methods. 
We broadly categorize these baselines into foundational approaches and recent SOTA architectures. 
The foundational group encompasses classic statistical and machine learning methods: VAR~\cite{Anderson1976TimeSeries2E}, LOF~\cite{10.1145/342009.335388}, OCSVM~\cite{Tax2004}; density estimation and autoregressive models: DAGMM~\cite{zong2018deep}, MMPCACD~\cite{7859276}, CL-MPPCA~\cite{10.1145/3292500.3330776}, LSTM~\cite{10.1145/3219819.3219845}; deep clustering and one-class networks: Deep-SVDD~\cite{pmlr-v80-ruff18a}, ITAD~\cite{10.1145/3340531.3412716}, THOC~\cite{NEURIPS2020_97e401a0}; and classic deep generative and reconstruction-based models: LSTM-VAE~\cite{8279425}, OmniAnomaly~\cite{10.1145/3292500.3330672}, BeatGAN~\cite{10.5555/3367471.3367658}, InterFusion~\cite{10.1145/3447548.3467075}. 
Furthermore, to demonstrate the superiority of POST against the latest advancements, our comparison includes recent SOTA deep models, which are mainly grouped into Transformer-based architectures: TranAD~\cite{10.14778/3514061.3514067}, AT~\cite{xu2022anomaly}, STAT~\cite{FU2024119978}; spatial-temporal GNNs: MST-GAT~\cite{DING2023527}, TopoGDN~\cite{10.1145/3627673.3679614}, LGAT~\cite{10.1016/j.neucom.2024.129024}; along with other advanced hybrid and generative frameworks: ImDiffusion~\cite{chen2023imdiffusion}, TSAD~\cite{10.1016/j.neucom.2025.130611} and CSCAD~\cite{10.1016/j.ipm.2025.104315}.
The detailed quantitative results across all five datasets are summarized in Table~\ref{tab:main_results}.

From the results one can observe that, where the F1-score is treated as the most crucial and comprehensive metric in the MTSAD literature, our proposed POST framework consistently and significantly outperforms all existing methods.
As reported in the final ``Avg'' column, POST achieves an average F1-score of 97.52\% across all five datasets, delivering a substantial absolute improvement of 2.61\% over the second-best method, AT (94.91\%).
This consistent superiority highlights the robustness of POST across diverse industrial domains.
Furthermore, a detailed examination of the Precision and Recall metrics reveals the underlying efficacy of our architectural design.
Conventional GNN-based spatial models often struggle with unconstrained reconstruction capabilities, forcing a compromise between Precision and Recall due to over-generalization.
In contrast, POST effectively mitigates this trade-off.
For instance, on the SWaT dataset, POST achieves a Recall of 100.00\% while simultaneously maintaining a remarkably high Precision of 96.08\%.
Similar balanced excellence is observed on the SMAP and PSM datasets, where F1-scores exceed 98\%.

As discussed in Section~\ref{sec:pre}, arbitrarily increasing model complexity is typically ineffective in MTSAD, as it tends to results in the over-generalization dilemma.
Our experimental observations mainly align with this analysis.
Specifically, POST retains the same depth of attention layers as the AT baseline, and only integrates the SAGA module with a marginal number of extra parameters.
The performance of POST convincingly demonstrates that the explicit modeling and adversarial optimization of spatial relationships, rather than brute-force scaling, are the key drivers of its superior performance.

\subsection{Spatial Anomaly Localization on SMD+}
\label{sec:exp_smd+}
To evaluate the spatial anomaly localization performance of various methods, we benchmark seven baseline models alongside our proposed POST on the SMD+ dataset in Table~\ref{tab:smd_plus_results}.
These baselines are categorized into three self-implemented conventional models (VAR~\cite{Anderson1976TimeSeries2E}, MMPCACD~\cite{7859276}, LSTM~\cite{10.1145/3219819.3219845}) and four SOTA deep models (AT~\cite{xu2022anomaly}, TranAD~\cite{10.14778/3514061.3514067}, ImDiffusion~\cite{chen2023imdiffusion}, TopoGDN~\cite{10.1145/3627673.3679614}). 
For all methods, the calculation of channel-wise metrics is identical to the time-wise evaluation protocol, except that the metric components are accumulated across all channels.
For TranAD, TopoGDN, and the three conventional methods, since their anomaly scores are inherently derived from channel-wise reconstruction errors, we directly evaluate their channel-wise metrics using their native 2D outputs.
Subsequently, their time-wise anomaly scores are computed by averaging the 2D scores as standard practice.
Conversely, AT utilizes Eq.~(\ref{eq:as_at}) to directly generate a time-wise score, and ImDiffusion relies on an ensemble voting mechanism. 
Both of them inherently lack native channel-wise outputs. 
Thus, we adopt their 1D time-wise scores and broadcast them to all channels to evaluate their spatial localization performance.
\begin{table}[htbp]
  \caption{Performance comparison of time-wise anomaly detection and channel-wise anomaly localization on the SMD+ benchmark. P, R, and F1 represent Precision, Recall, and F1-score (in \%), respectively.}
  \label{tab:smd_plus_results}
  \centering  
  \resizebox{\columnwidth}{!}{  
  \renewcommand{\arraystretch}{1.3}
  \begin{tabular}{lcccccc}                        
  \toprule                                    
  \multicolumn{1}{c}{\multirow{3}*{Method}} & \multicolumn{3}{c}{Time-wise} & \multicolumn{3}{c}{Channel-wise} \cr     
  \cmidrule{2-7}             
   & P & R & F1 & P & R & F1 \\
  \midrule
  VAR~\cite{Anderson1976TimeSeries2E} & \textbf{93.50} & 34.64 & 50.56 & 32.63 & 52.88 & 40.36 \\
  MMPCACD~\cite{7859276} & 63.69 & 20.29 & 30.78 & 21.81 & 35.78 & 27.10 \\
  LSTM~\cite{10.1145/3219819.3219845} & \underline{92.32} & 28.06 & 43.04 & 47.03 & 57.28 & 51.65 \\
  AT~\cite{xu2022anomaly} &  86.87 & 73.19 & 79.44 & 28.62 & \underline{67.04} & 40.12 \\
  TranAD~\cite{10.14778/3514061.3514067}  & 84.60 & 75.79 & 79.03 & \textbf{51.70} & 60.89 & 55.53 \\
  ImDiffusion~\cite{chen2023imdiffusion}  & 83.05 & \textbf{82.39} & \underline{82.88} & 31.82 & 53.79 & 41.50 \\
  TopoGDN~\cite{10.1145/3627673.3679614} & 84.31 & 74.09 & 78.56 & 50.12 & 64.69 & \underline{56.09} \\
  \midrule
  POST (Ours) & 88.68 & \underline{80.10} & \textbf{84.17} & \underline{50.97} & \textbf{73.22} & \textbf{60.10} \\
  \bottomrule
  \end{tabular} 
  }
\end{table}

From the performance comparison, we can observe that for the challenging SMD+ benchmark, conventional methods adopt a highly conservative strategy.
They only trigger alarms when encountering severe anomalies.
This results in remarkably high Precision but abysmal Recall, indicating severe under-reporting. Conversely, SOTA deep models establish a clear advantage in the time-wise anomaly detection task, achieving a substantial F1-score improvement ($>30\%$) over traditional baselines by more accurately distinguishing anomalies. 
However, a critical limitation emerges in models like AT and ImDiffusion. 
Their dedicated output designs effectively enchance the performance in time-wise detection.
But when broadcasting their 1D scores to all channels, an overwhelming number of normal channels are falsely labeled as anomalous, leading to a catastrophic drop in Precision (e.g., 28.62\% for AT) and yielding channel-wise F1 scores that are indistinguishable from conventional methods. 
Especially for LSTM, by simply relying on native channel-wise reconstruction errors, it even slightly outperforms these 1D-optimized SOTA models in spatial anomaly localization.

For the remaining SOTA methods (TranAD and TopoGDN), the employment of discriminative deep architectures combined with native channel-wise reconstruction errors enables them to achieve competitive results in both tasks. 
Nevertheless, our proposed POST maintains a notable advantage across both evaluations, outperforming them by over 4\% in F1-score. 
This superiority verifies the efficacy of incorporating spatio-temporal association discrepancy into the final anomaly score. 
By forcing the model to evaluate anomalies not merely on signal reconstruction, but also on the deviations in temporal dependencies and spatial topological correlations, POST prevents the model from simply memorizing all inputs to mask anomalies.
It is particularly crucial as modern reconstruction models quickly grow larger.
This observation also aligns with the results in~\cite{xu2022anomaly}. 
Finally, by explicitly introducing the spatial discrepancy in Eq.~(\ref{eq:asts}), POST leads the spatial anomaly localization task by a significant margin of 4.01\% in F1-score over the second-best model (TopoGDN). 
Note that since no clear improvement is observed when adopting Eq.~(\ref{eq:asts}) in standard time-wise anomaly detection benchmarks, we choose to retain Eq.~(\ref{eq:as_at}) as the standard output in other comparisons.

\subsection{Verification of Proposed Components}
To rigorously validate the efficacy of the core mechanisms proposed in POST, we conduct a comprehensive component analysis on the SMD dataset. 
As reported in Table~\ref{tab:ablation_smd}, the evaluation systematically ablates the architectural designs and the core adversarial learning mechanism.
The evaluated variants are categorized into the following two groups: 

1) \textit{Temporal Association Refinement (AT$\rightarrow$TASA):} This group demonstrates the necessity of our proposed DPE when explicit spatial topology is introduced, transitioning from the original attention in AT to the TASA module.
\begin{itemize}
    \item AT$-$APE: The AT baseline stripped of its standard APE.
    \item AT$-$APE$+$SAGA: Integrates SAGA into the baseline without applying any positional encoding.
    \item AT$+$SAGA: Directly plugs the SAGA module into AT with its original entangled positional encoding.
    \item AT$-$APE$+$RoPE: Substitutes the APE with RoPE~\cite{su2023roformerenhancedtransformerrotary}, another typical positional encoding method which can be disentangled within the attention module.
    \item AT$-$APE$+$MLP-DPE: Maintains our proposed DPE but replaces the analytical distance matrix with a fully learnable multi-layer perceptron (MLP).
\end{itemize}

2) \textit{SAGA Module (TASA$+$SAGA):} This group isolates the contributions of the learnable spatial topology and adversarial learning mechanism of the SAGA module within the full TASA$+$SAGA framework.
\begin{itemize}
    \item w/o SAGA: Removes the spatial module, relying solely on TASA modules.
    \item Fixed $\mathcal{G}^l$: Freezes the adjacency matrices after initialization in SAGA, eliminating the dynamic topological evolution during training.
    \item w/o $\text{AssDis}_s$: Retains the full SAGA architecture but removes the prior-observation adversarial optimization of $\text{AssDis}_s$, updating the spatial parameters relying only on the reconstruction loss and structural regularizations in Algorithm~\ref{alg:train}.
    \item Identity Init of $\mathcal{G}^l$: Initializes $\mathcal{G}^l$ with an identity matrix rather than the $k$NN structural prior.
\end{itemize}

\begin{table}[htbp]
  \centering
  \caption{Component analysis of POST on the SMD dataset. 
  Precision, Recall, and F1-score are reported in \%.}
  \label{tab:ablation_smd}
  \resizebox{\columnwidth}{!}{%
  \begin{tabular}{lccc}
    \toprule
    Variant & Precision & Recall & F1-score \\
    \midrule
    
    \multicolumn{4}{l}{\textit{Baseline Model}} \\
    AT~\cite{xu2022anomaly} & 89.40 & 95.45 & 92.33 \\
    \midrule  

    \multicolumn{4}{l}{\textit{AT$\rightarrow$TASA}} \\
    AT$-$APE & 90.14 & 88.66 & 89.87 \\
    AT$-$APE$+$SAGA & 90.32 & 90.86 & 90.50 \\
    AT$+$SAGA & 88.96 & 87.26 & 88.17 \\
    AT$-$APE$+$RoPE & 91.65 & 92.08 & 91.86 \\
    AT$-$APE$+$MLP-DPE & 91.25 & 93.16 & 92.19 \\
    \midrule
    
    \multicolumn{4}{l}{\textit{TASA$+$SAGA}} \\
    w/o SAGA & 89.18 & 95.61 & 92.58 \\
    Fixed $\mathcal{G}^l$ & 91.88 & 95.79 & 93.84 \\
    w/o $\text{AssDis}_s$ & 91.45 & 97.16 & 94.22 \\
    Identity Init of $\mathcal{G}^l$ & 94.16 & 96.23 & 95.19 \\
    \midrule
    
    POST (Full Model) & \textbf{94.48} & \textbf{97.23} & \textbf{95.83} \\
    \bottomrule
  \end{tabular}%
  }
\end{table}
\begin{figure*}[t]
    \centering
    \includegraphics[width=\textwidth]{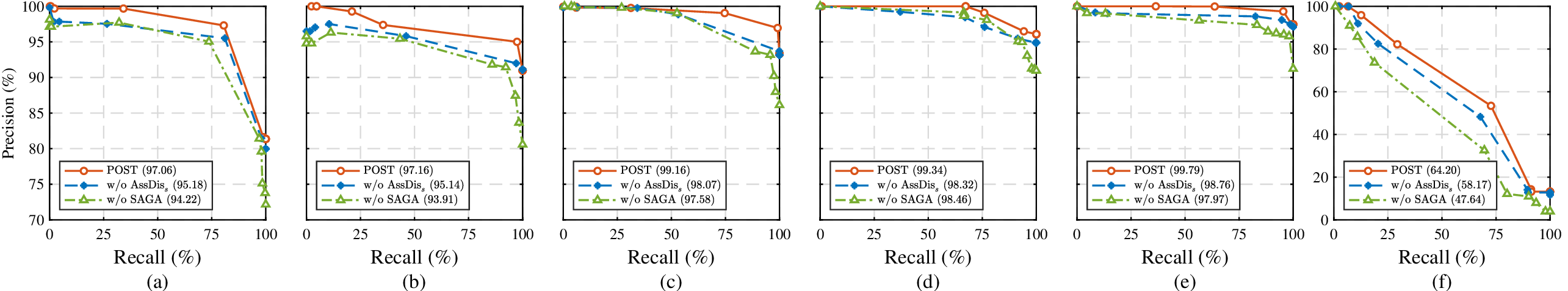}
    \caption{Precision-recall curves of different configurations across six datasets: (a) SMD, (b) MSL, (c) SMAP, (d) SWaT, (e) PSM, and (f) SMD+. 
    Our proposed POST and its two variants (w/o SAGA and w/o $\text{AssDis}_s$) from Table~\ref{tab:ablation_smd} are compared here.
    The operating points on each curve are dynamically generated by varying the expected anomaly ratio threshold from $10^{-5}\%$ to $30.0\%$ to compute the corresponding score percentiles.
    We also report the average precision (AP) after each configuration as a percentage in parentheses within the legend.
    The evaluation on the SMD+ dataset is conducted under the channel-wise protocol detailed in Section~\ref{sec:exp_smd+}, whereas the results on the other five datasets are evaluated following the standard MTSAD protocol.}
    \label{fig:pr_curves}
\end{figure*}

The first group of comparisons in Table~\ref{tab:ablation_smd} reveals the impact of PE on temporal association capabilities, particularly concerning the Recall metric. 
Removing APE mechanisms from the baseline (AT$-$APE) fundamentally results in a severe Recall degradation ($>6\%$). 
Theoretically, in the original AT framework, the adversarial learning around $\text{AssDis}_t(\mathcal{P},\mathcal{S})$ relies on the temporal positional information of the input to guide $\mathcal{S}$ in effectively diverging from the peak of the Gaussian prior $\mathcal{P}$.
The absence of PE causes this adversarial process to collapse, yielding unreliable association discrepancies that fail to distinguish anomalies from the normal temporal context, ultimately leading to significant false negatives. 
Furthermore, while integrating the SAGA module without PE (AT$-$APE$+$SAGA) provides marginal gains, it remains insufficient to compensate for the loss of temporal awareness. 
Directly plugging the SAGA module into the standard AT (AT$+$SAGA) leads to catastrophic consequences. 
Due to the noise brought by PE in the input, SAGA exerts a negative influence on signal reconstruction from a spatial perspective, resulting in the worst F1-score across all variants. 
By implementing the proposed DPE, we successfully decouple the temporal position variables from the spatial topology and maintain a foundational model (w/o SAGA) with comparable performance to the original AT for subsequent spatial augmentations.
Here, we also add two PE variants as comparison items.
Both methods implement PE internally within the attention module.
However, it can be observed that the RoPE scheme (AT$-$APE$+$RoPE) is unsuitable for MTSAD scenarios, which feature fixed and relatively short sliding window lengths.
Also, incorporating an MLP into the DPE (AT$-$APE$+$MLP-DPE) yields no clear performance improvement.
Thus we omit these extra learnable parameters from the final design.

From Table~\ref{tab:ablation_smd}, one can also observe that building upon the TASA module, the introduction of SAGA primarily contributes to the continuous enhancement of Precision across its variants. 
The adversarial learning around $\text{AssDis}_s(\hat{\mathcal{G}},\mathcal{A})$ plays a pivotal role in this process. 
A comparison between Fixed $\mathcal{G}^l$ and w/o $\text{AssDis}_s$ shows that optimizing the $\mathcal{G}^l$ based solely on signal reconstruction yields insufficient performance, with Precision remaining below $92\%$. 
However, activating the minimax adversarial learning forces the model to gain a Precision over $94\%$.
The adversarial learning within SAGA assists $\mathcal{G}^l$ in better tracking the statistical characteristics of the observation across the training set, finally leading to a more reasonable posterior $\tilde{\mathcal{A}}^l$ in Eq.~(\ref{eq:saga_a_posterior}). 
With the reconstruction from a spatial perspective, the signal processed by SAGA exhibits clearer channel-wise decoupling, which facilitates more effective analysis by the subsequent TASA module. 
Moreover, note that the $k$NN-based initialization leverages the statistical features of the raw training data to provide an ideal starting point for the optimization of $\mathcal{G}^l$, yielding stable performance improvements. 
As a comprehensive integration of above components, our final proposed POST model achieves a superior balance of $94.48\%$ Precision and $97.23\%$ Recall, validating the efficacy of minimax-optimized association modeling in both temporal and spatial dimensions.
\begin{figure*}[t]
  \centering
  \includegraphics[width=\textwidth]{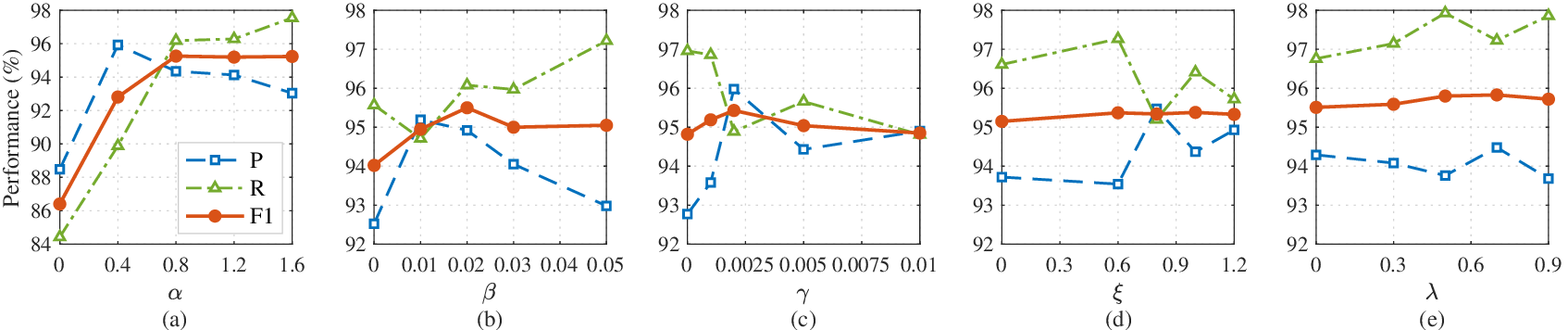}
  \caption{Sensitivity analysis of hyperparameters on the SMD dataset. Performance is evaluated using Precision (P), Recall (R), and F1-score (in \%). 
  We adopt a two-phase search strategy. (a)-(d) illustrate the orthogonal search process for the loss weights $\alpha, \beta, \gamma$, and $\xi$ with $\lambda=0$. 
  We first establish a baseline configuration ($\alpha=1.0, \beta=0.01, \gamma=0.001$ and $\xi=1.0$) by tracking their initial gradient magnitudes during training, and then vary each parameter individually while keeping the others fixed. 
  (e) presents the independent search for the proximal step size $\lambda$, performed after the optimal combination of the aforementioned weights is identified.}
  \label{fig:hyper_sensitivity}
\end{figure*}

\textbf{Analysis of Decision Boundary}: 
To intuitively illustrate the influence of the proposed mechanisms in SAGA across varying decision boundaries, Fig.~\ref{fig:pr_curves} visualizes the precision-recall (PR) curves for POST and its two key variants across six datasets. 
From the comparison, it can be observed that the full POST model consistently achieves the largest average precision (AP) in all scenarios, demonstrating a superior capability to maintain high precision across varying recall levels and finally yielding the highest F1-scores. 
In contrast, when the spatial modeling is entirely discarded (w/o SAGA), the PR curves exhibit premature and steep declines in the early stages. 
A slight relaxation of the anomaly threshold immediately triggers a surge in false positives. 
Furthermore, there exists a long-tail effect in the later stages of these curves, implying that a substantial number of anomalies are assigned extremely low scores.
Building upon TASA, incorporating SAGA without adversarial optimization (w/o $\text{AssDis}_s$) effectively mitigates these severe false negatives. 
Compared to the full POST model, this variant only shows a precision gap in the early stages of the curves.

We also evaluate the models under the channel-wise anomaly localization protocol on the SMD+ dataset as depicted in Fig.~\ref{fig:pr_curves}(f). 
Since the w/o SAGA variant lacks spatial modeling mechanisms, it cannot generate native channel-wise anomaly scores. 
We adopt the same broadcast strategy as AT in Section~\ref{sec:exp_smd+} to evaluate its performance. 
This fundamental limitation results in a massive $16.56\%$ AP drop compared to the full POST model and triggers a catastrophic precision degradation in the early recall stages. 
On the other hand, the variant w/o $\text{AssDis}_s$ retains the SAGA module and can compute channel-wise scores directly.
However, the absence of adversarial learning still renders the generated association discrepancies insufficiently reliable for more challenging anomaly localization tasks.

\subsection{Hyperparameter Analysis}
\label{sec:hyp_param}
The optimization of POST involves several key hyperparameters that govern the balance between reconstruction quality and structural regularizations. 
To identify the optimal configuration, we adopt a two-phase search strategy. 
As detailed in Fig.~\ref{fig:hyper_sensitivity}, we first search the direct coefficients for four loss terms in Algorithm~\ref{alg:train}: $\text{AssDis}_t, \text{AssDis}_s, \mathcal{L}_s$, and $\mathcal{L}_{\text{tr}}$. 
During this process, the $\ell_1$ sparsity constraint on the adjacency matrix in Eq.~(\ref{eq:prox_iter}) is omitted ($\lambda=0$). 
Through an initial magnitude alignment with the reconstruction loss $\mathcal{L}_{\text{rec}}$, we establish an approximate baseline configuration for these four parameters. 
Then we adjust each parameter individually to observe the corresponding performance variations.

As illustrated in the figure, the F1-score serves as a highly reliable comprehensive metric. 
Evaluated by this metric, the temporal association modeling centered around $\text{AssDis}_t$ remains the core driving force of the entire model. 
Its absence results in a severe performance degradation. 
However, when $\alpha > 0.8$, the performance tends to stabilize. 
This trend is largely consistent with observations in \cite{xu2022anomaly}, except the specific values differ slightly due to additional loss terms. 
Building upon this, the weights of the spatial constraints, $\beta$ and $\gamma$, exhibit higher sensitivity. 
Values that are either too large or too small lead to performance degradation, indicating that a reasonable balance is required between learning spatial associations and regularization of $\mathcal{G}$. 
In contrast, $\xi$ demonstrates strong robustness over a wide range, providing a stable constraint for $\mathcal{S}$.

In the second phase, we choose the optimal configuration of the aforementioned weights and conduct a search for the step size parameter $\lambda$ in the proximal step.
As indicated in Algorithm~\ref{alg:train}, this parameter independently regularizes $\mathcal{G}$ after the completion of other gradient descent iterations, which justifies our choice to search it separately.
As shown in Fig.~\ref{fig:hyper_sensitivity}(e), the application of $\lambda$ provides a final performance boost, elevating the F1-score to the global optimum.
Ultimately, we identify the optimal configuration of all hyperparameters as listed in Section~\ref{sec:impl} for all other experiments.

\subsection{Visualization of Associations}
\begin{figure}[t]
    \centering
    \includegraphics[width=\columnwidth]{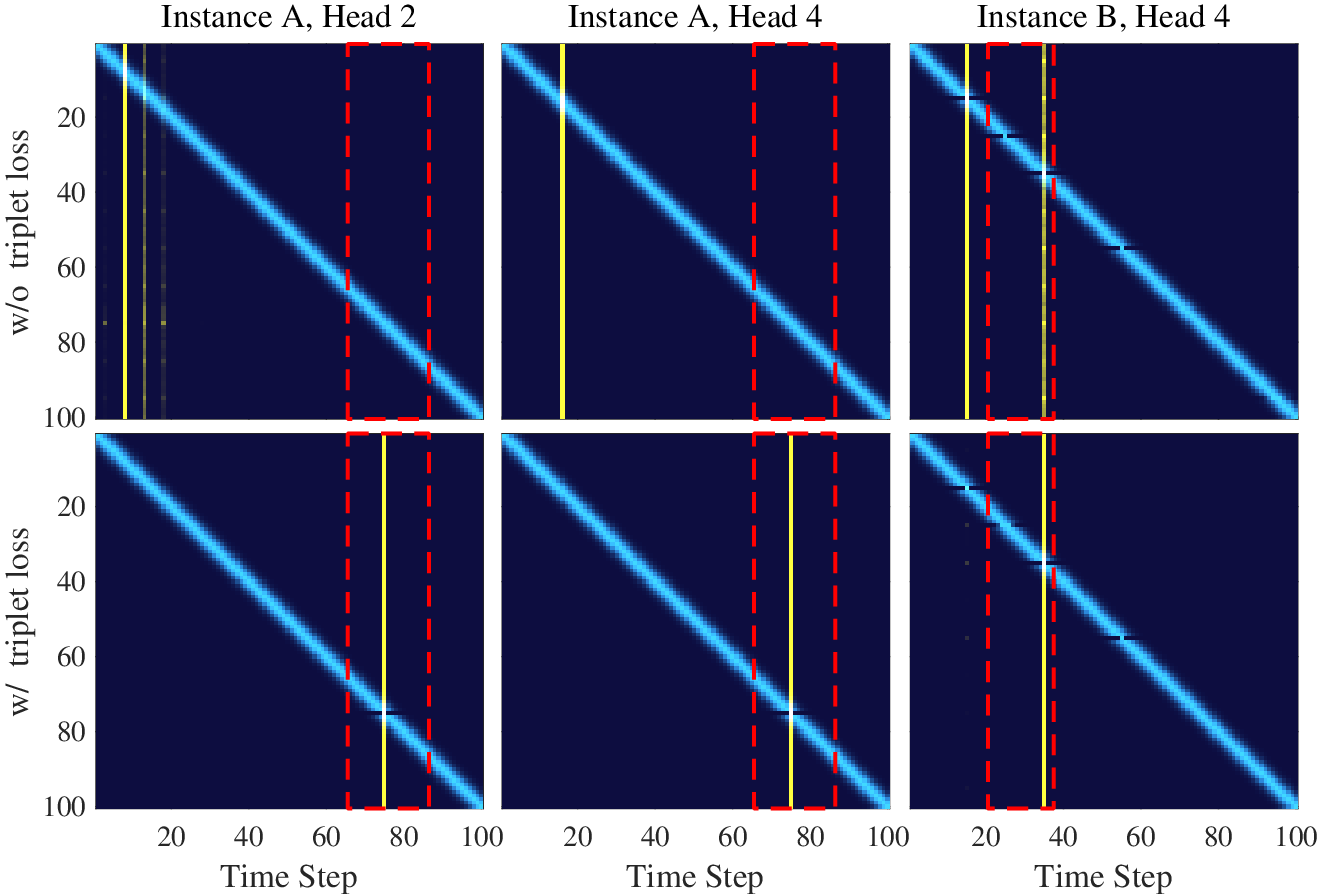}
    \caption{Visualization of the learned temporal association $\mathcal{S}$ and the Gaussian prior $\mathcal{P}$.
    The heatmaps are generated by overlaying $\mathcal{S}$ and $\mathcal{P}$ extracted from various attention heads within the TASA module during the inference phase.
    The light blue diagonal lines trace the peaks of $\mathcal{P}$.
    As defined in Eq.~(\ref{eq:p_gauss}), the peak values of $\mathcal{P}$ are localized at each time step.
    The bright yellow regions highlight the peak attention scores of $\mathcal{S}$.
    The red dashed boxes indicate the ground-truth anomalous regions. 
    To illustrate cross-instance variations, the first two columns display different heads from the same input instance (Instance A), whereas the third column visualizes a head from a distinct instance (Instance B). 
    } 
    \label{fig:attention_vis}
\end{figure}

In this subsection, we conduct a visul analysis to intuitively demonstrate the influence of the proposed regularization terms to the temporal and spatial associations.
First, to investigate the impact of the proposed regularization $\mathcal{L}_{\text{tr}}$ in Eq.~(\ref{eq:reg_tr}) to the TASA module, we visualize the learned temporal association $\mathcal{S}$ and the Gaussian prior $\mathcal{P}$ in Fig.~\ref{fig:attention_vis}. 
Essentially, to maximize the divergence between $\mathcal{S}$ and $\mathcal{P}$ during the adversarial phase, the model learns to aggressively concentrate the attention scores into sharp peaks. 
However, as illustrated in the top row of Fig.~\ref{fig:attention_vis}, optimizing this objective without $\mathcal{L}_{\text{tr}}$ leads to a degenerate trivial solution. 
The model tends to permanently anchor the peaks to static temporal positions, disregarding the actual input signals. 
Consequently, the generated discrepancy $\text{AssDis}_t$ loses its capacity to discriminate anomalies.
In contrast, under the constraint of $\mathcal{L}_{\text{tr}}$, the model is compelled to dynamically link the peaks of $\mathcal{S}$ with the actual input features.
As demonstrated in the bottom row of Fig.~\ref{fig:attention_vis}, the resulting peaks of $\mathcal{S}$ strike exactly at those collapsed moments of $\mathcal{P}$, better aligning with the original theoretical premise of $\text{AssDis}_t$. 

\begin{figure}[t]
    \centering
    \includegraphics[width=\columnwidth]{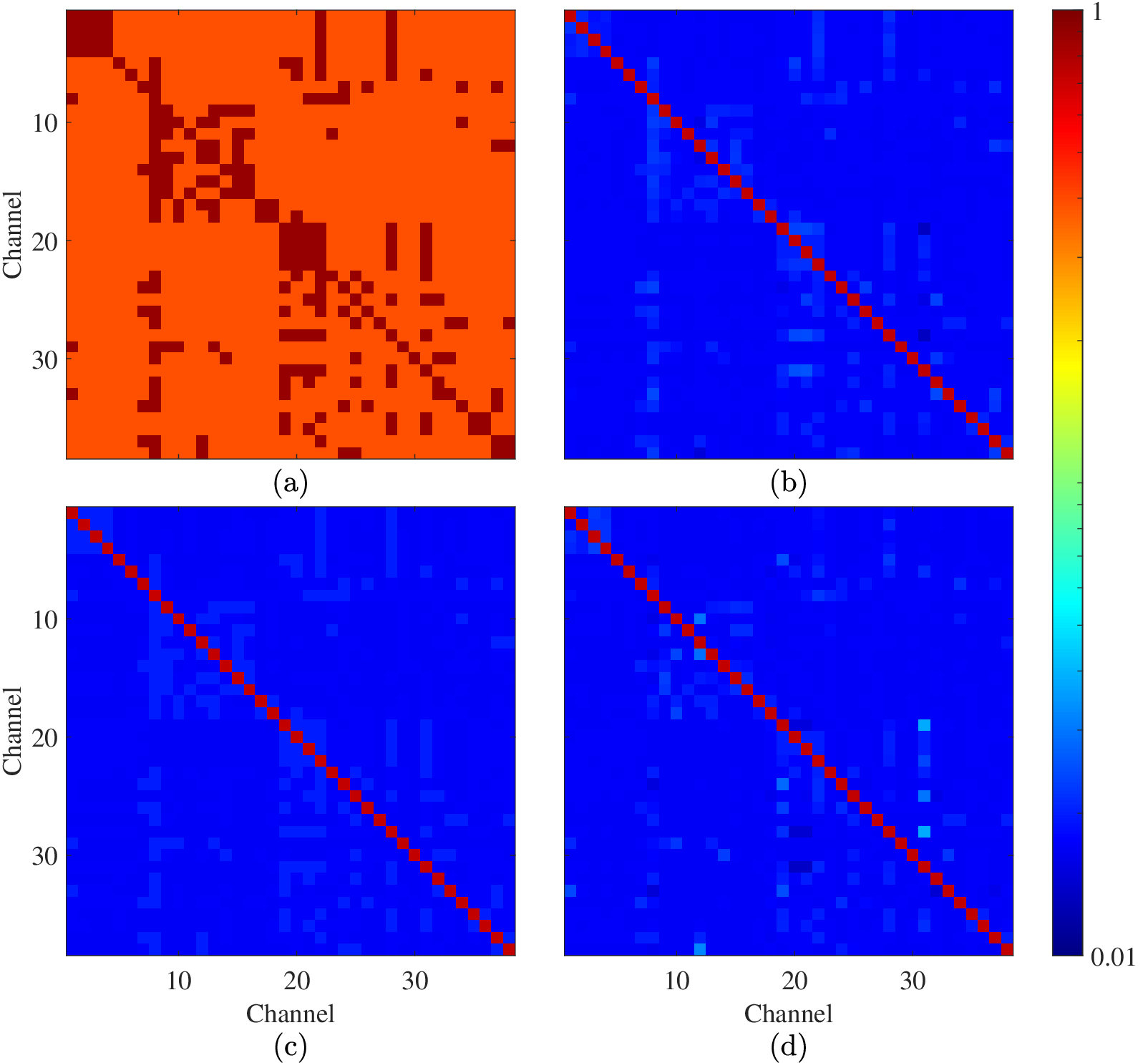}
    \caption{Visualization of the spatial adjacency matrices $\mathcal{G}^l$ across different network layers. 
    Here (a) depicts the initial prior graph constructed via the $k$NN algorithm.
    (b), (c), and (d) illustrate the learned spatial graphs at Layer 1-3, respectively. 
    The color intensity indicates the magnitude of the spatial association weights between different channels.
    To facilitate intuitive visual comparison, a Sigmoid activation is applied to uniformly map the dynamic range of all heatmaps to $(0, 1)$.
    } 
    \label{fig:g_vis}
\end{figure}

Following the temporal analysis, we also investigate the influence of structural regularization on the spatial adjacency matrices $\mathcal{G}^l$. 
In Fig.~\ref{fig:g_vis}, we present the learned heatmaps of $\mathcal{G}^l$ within SAGA modules at different layers, alongside the initial prior graph constructed based on the statistical characteristics of the training set. 
Compared to the static initialization, the synergy of reconstruction and adversarial learning produces more targeted spatial priors that adaptively reflect the intrinsic data features of each layer. 
Notably, with the guidance of smoothness constraint $\mathcal{L}_s$ and the $\ell_1$ sparsity constraint, the model exhibits a higher proportion of self-association along the diagonal, while effectively suppressing unreliable cross-channel correlations. 
Furthermore, an evolutionary trend can be observed across Layers 1-3 in Fig.~\ref{fig:g_vis}(b)-(d). 
In the first two layers, the model primarily constrains spatial associations to induce sparsity, filtering out potential inter-channel noise. 
In the third layer, the model begins to enhance reliable spatial dependencies, facilitating more effective signal reconstruction.

\subsection{Sparsity Constraint on $\tilde{\mathcal{G}}^l$}
\label{sec:exp_l1}
\begin{figure}[t]
  \centering
  \includegraphics[width=1.0\linewidth]{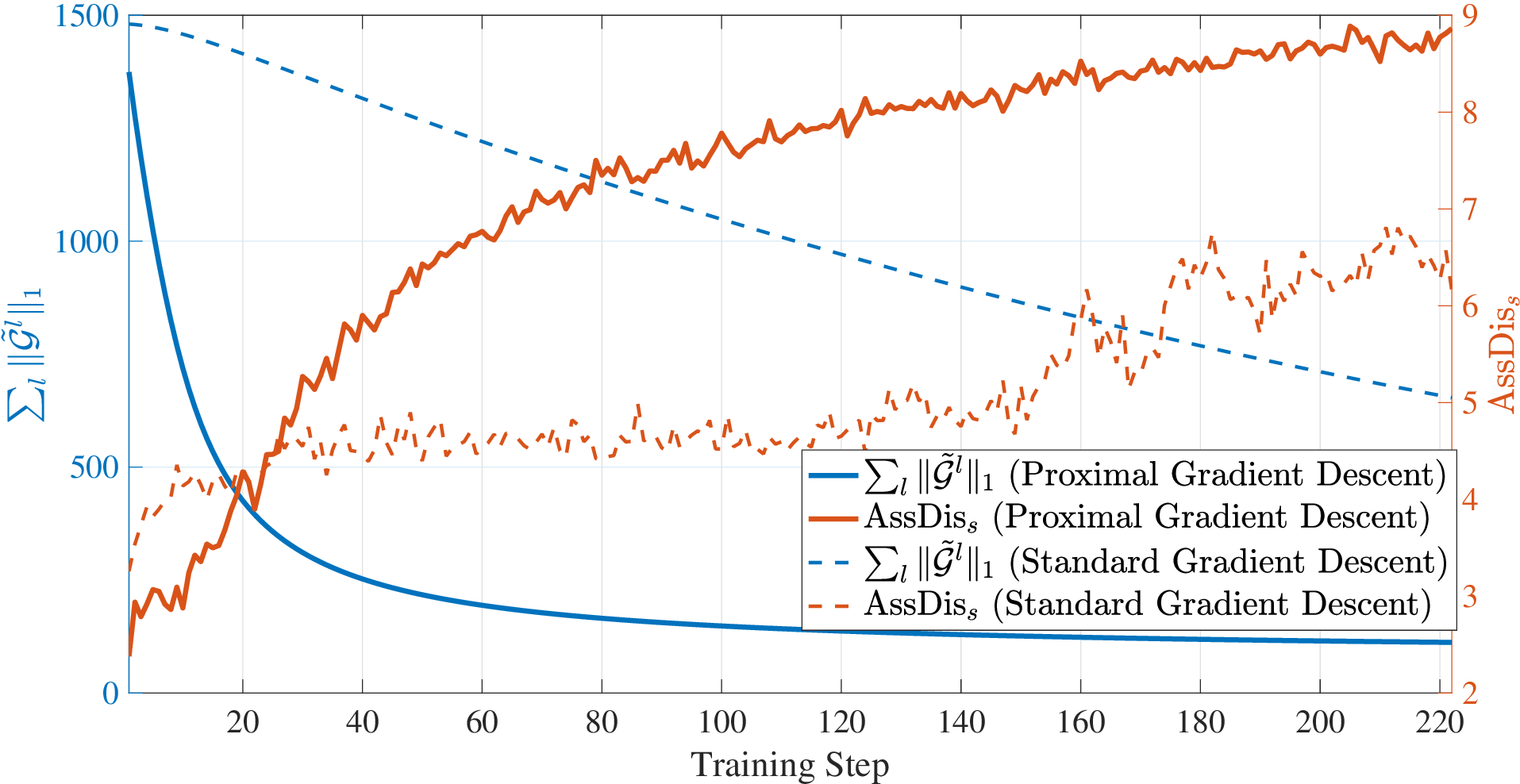}
  \caption{Optimization dynamics of the spatial topology during learning.
  The left axis (blue) tracks the $\ell_1$ norm of $\tilde{\mathcal{G}}^l$.
  The right axis (orange) tracks the spatial association discrepancy ($\text{AssDis}_s$).
  Solid lines indicate the proposed proximal gradient descent scheme, while dashed lines represent the standard gradient descent baseline.
  }
  \label{fig:saga_dynamics}
\end{figure}

To study the impact of different sparsity constraint schemes aforementioned at the end of Section~\ref{sec:saga}, we compare the optimization dynamics of the $\ell_1$ norm of $\tilde{\mathcal{G}}^l$ and $\text{AssDis}_s$ in Fig.~\ref{fig:saga_dynamics}.
Here, we evaluate our proposed proximal gradient descent scheme against a standard gradient descent baseline.
The baseline directly incorporates $\sum_{l}\|\tilde{\mathcal{G}}^l\|_1$ as a regularization term into the overall loss, updating $\mathcal{G}^l$ via standard backpropagation (Step 4 of Algorithm~\ref{alg:train}).
As illustrated, by applying the proximal operator in Eq.~(\ref{eq:prox_iter}), multiple fixed-point iterations at each training step effectively ensure the sparsity of $\mathcal{G}^l$, thereby highlighting the genuinely statistically relevant channel associations.
On top of that, the adversarial learning also effectively enlarges $\text{AssDis}_s$ on normal signals, converging to a substantially higher plateau.
This amplified discrepancy on normal observations guarantees the discriminative capability for anomaly detection, ultimately improving the overall model performance.

\subsection{Case Study of Spatio-Temporal Anomaly Localization}
\label{sec:exp_atst}
\begin{figure}[t]
  \centering
  \includegraphics[width=1.0\linewidth]{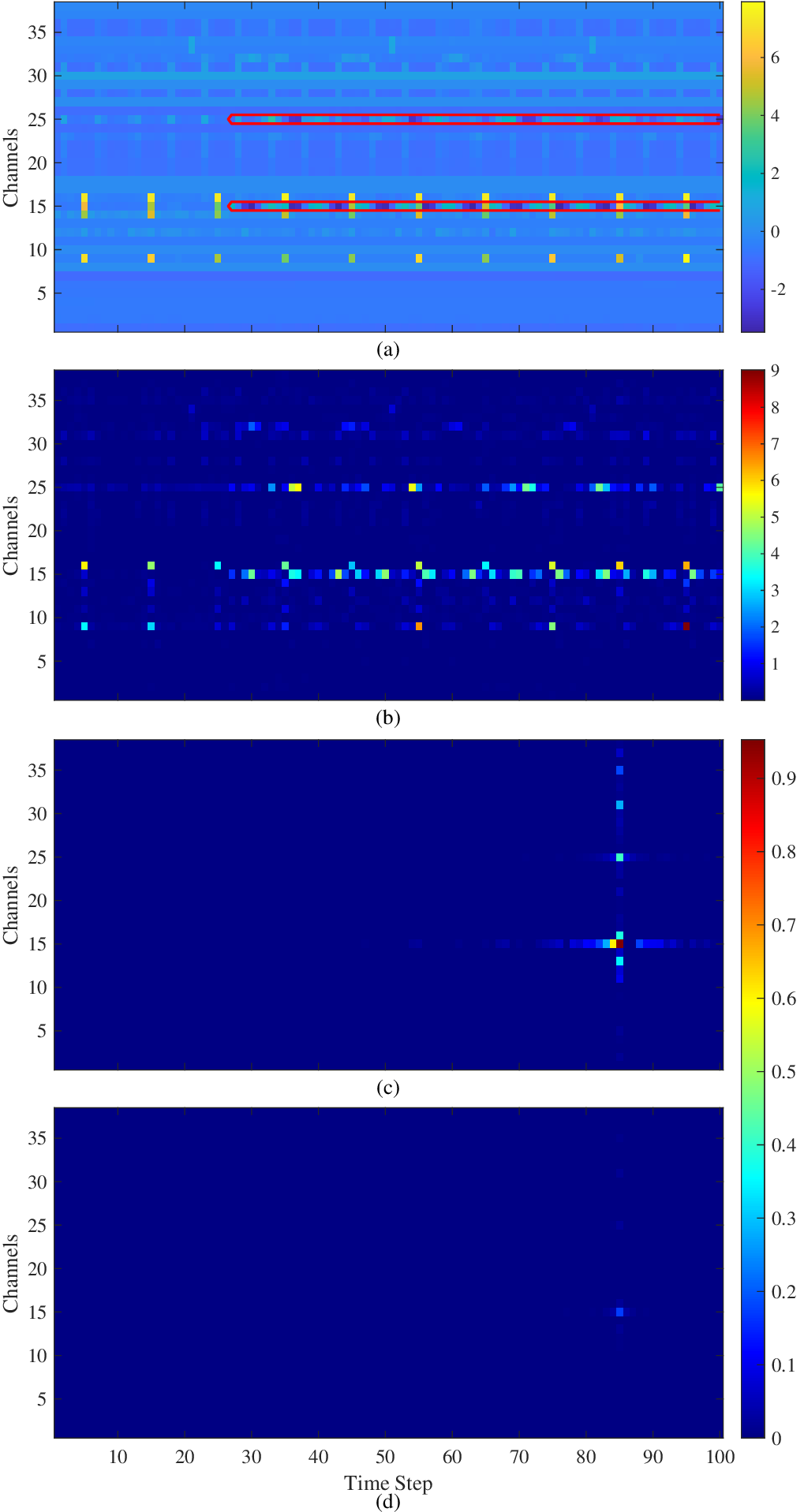}
  \caption{Heatmap visualization of the spatial anomaly localization on the SMD+ dataset.
  We take a 100-step time window containing a collective anomaly as an instance, illustrating the entire process from raw inputs to the final anomaly scores. 
  (a) Raw signals with ground-truth channel-wise anomalies indicated by red bounding boxes.
  (b) Pure reconstruction error between the input and output of POST.
  (c) The final anomaly score $\text{AS}_{ts}$ calculated via Eq.~(\ref{eq:asts}).
  (d) A variant of (c) where the Sigmoid activation for $\text{AssDis}_s$ is replaced with Softmax.
  To ensure an intuitive visual comparison, (c) and (d) are mapped to a unified dynamic range and share the same colorbar.
  }
  \label{fig:vis_atst}
\end{figure}

To explicitly validate the design of the proposed spatio-temporal anomaly score $\text{AS}_{ts}$ in Eq.~(\ref{eq:asts}), we present an anomaly detection case from the SMD+ dataset in Fig.~\ref{fig:vis_atst}.
As depicted in Fig.~\ref{fig:vis_atst}(a), the sample exhibits continuous anomalies on two specific channels, which are relatively sparse in the spatial dimension.
For this challenging scenario, as shown in Fig.~\ref{fig:vis_atst}(b), relying solely on the pure reconstruction error to identify anomalies introduces severe noise.
Some scores at incorrect temporal and spatial positions even exceed those at the actual anomalous locations.
In contrast, the integration of reconstruction error, $\text{AssDis}_t$, and $\text{AssDis}_s$ into the final anomaly score effectively suppresses these noise.

Furthermore, to analyze the contribution of $\text{AssDis}_s$ in Eq.~(\ref{eq:asts}), we compare two different activation functions: our proposed Sigmoid function (Fig.~\ref{fig:vis_atst}(c)) and the conventional Softmax function (Fig.~\ref{fig:vis_atst}(d)) which is consistent with the design of $\text{AssDis}_t$ as in \cite{xu2022anomaly}.
From the comparison, it can be observed that the Softmax activation suffers from a severe masking effect.
Since Softmax enforces a competitive normalization, it drastically compresses the overall magnitude of the anomaly scores.
When mapped to a unified dynamic range, the anomalous points in Fig.~\ref{fig:vis_atst}(d) become almost indistinguishable, indicating a critical loss of discriminative capability.
In standard time-wise anomaly detection, this issue may be mitigated by the inherent temporal continuity of anomalies and the point-adjustment mechanism in post-processing.
However, for the spatial anomaly localization task, channel-wise anomalies are spatially discrete and sparse.
In this case, it is more reasonable to evaluate each channel independently with Sigmoid activation.
As evidenced by Fig.~\ref{fig:vis_atst}(c), this asymmetric activation scheme ensures a precise and robust capture of spatio-temporal anomalies. 

\section{Conclusion}
In this paper, we proposed POST, a novel unsupervised framework for MTSAD.
POST conducts the prior-observation adversarial learning acorss both spatial and temporal dimensions within a unified framework, and integrates the resulting association discrepancies into the final anomaly score as auxiliary constraints to the reconstruction error for enhanced discriminative capacity.
With the channel-wise anomaly criterion of POST, we discuss the problem of anomaly localization in MTS, proposing the synthetic SMD+ benchmark with precise channel-wise annotations.
Extensive experiments across multiple datasets demonstrate that POST significantly outperforms existing SOTA methods in both standard time-wise detection and spatial anomaly localization tasks.
Future work will explore the dynamic spatio-temporal evolution of real-world faults to further enhance root-cause attribution capabilities of models.

\bibliographystyle{IEEEtran}
\bibliography{reference.bib}


 





\end{document}